\documentclass[11pt]{article}

\usepackage[final]{acl}

\usepackage{times}
\usepackage{latexsym}

\usepackage[T1]{fontenc}

\usepackage[utf8]{inputenc}

\usepackage{microtype}

\usepackage{inconsolata}

\usepackage{graphicx}
\usepackage{capt-of}
\usepackage{amsmath}
\usepackage{amssymb}
\usepackage{booktabs}
\usepackage{colortbl}
\usepackage{multirow}
\usepackage{algorithm}
\usepackage{algorithmic}
\usepackage{pifont}
\usepackage{amsthm}
\usepackage{enumitem}
\theoremstyle{definition}
\newtheorem{proposition}{Proposition}
\usepackage{titletoc}
\definecolor{AppTocBlue}{HTML}{2563EB}

\definecolor{RkBlueA}{HTML}{F7FAFD}
\definecolor{RkBlueB}{HTML}{D3E4F4}
\definecolor{RkBlueC}{HTML}{BDD7EE}
\definecolor{RkBlueD}{HTML}{9EC3E6}
\definecolor{RkGreenA}{HTML}{F8FCF6}
\definecolor{RkGreenB}{HTML}{D0E6C1}
\definecolor{RkGreenC}{HTML}{C6E0B4}
\definecolor{RkGreenD}{HTML}{A8D08D}
\definecolor{RkPeachA}{HTML}{FFF9F5}
\definecolor{RkPeachB}{HTML}{FCE3D3}
\definecolor{RkPeachC}{HTML}{F8CBAD}
\definecolor{RkPeachD}{HTML}{F4B183}
\definecolor{RkSlateA}{HTML}{F3F5F8}
\definecolor{RkSlateB}{HTML}{E2E8F0}
\definecolor{RkSlateC}{HTML}{CBD5E1}
\definecolor{RkSlateD}{HTML}{A8B8CC}
\definecolor{RkVioletA}{HTML}{F8F6FC}
\definecolor{RkVioletB}{HTML}{E0D5EE}
\definecolor{RkVioletC}{HTML}{D1C1E5}
\definecolor{RkVioletD}{HTML}{BDA5DA}
\definecolor{RkTealA}{HTML}{F3FBFA}
\definecolor{RkTealB}{HTML}{D0EBE7}
\definecolor{RkTealC}{HTML}{ACD9D2}
\definecolor{RkTealD}{HTML}{85C7BC}
\definecolor{RkRoseA}{HTML}{FFF6F8}
\definecolor{RkRoseB}{HTML}{F8DCE5}
\definecolor{RkRoseC}{HTML}{F1BFD0}
\definecolor{RkRoseD}{HTML}{E8A0B9}
\definecolor{RkAmberA}{HTML}{FFFBF0}
\definecolor{RkAmberB}{HTML}{FBEAC1}
\definecolor{RkAmberC}{HTML}{F5D993}
\definecolor{RkAmberD}{HTML}{EFC56C}
\definecolor{RkIndigoA}{HTML}{F6F7FC}
\definecolor{RkIndigoB}{HTML}{DFE3F4}
\definecolor{RkIndigoC}{HTML}{C6CFEC}
\definecolor{RkIndigoD}{HTML}{ADBCE3}
\newcommand{\ffRank}[2]{\cellcolor{RkBlue#1}#2}
\newcommand{\celebRank}[2]{\cellcolor{RkGreen#1}#2}
\newcommand{\dfdcRank}[2]{\cellcolor{RkPeach#1}#2}
\newcommand{\wildRank}[2]{\cellcolor{RkSlate#1}#2}
\newcommand{\deeperRank}[2]{\cellcolor{RkViolet#1}#2}
\newcommand{\forgeryRank}[2]{\cellcolor{RkTeal#1}#2}
\newcommand{\celebPlusRank}[2]{\cellcolor{RkRose#1}#2}
\newcommand{\dfFortyRank}[2]{\cellcolor{RkAmber#1}#2}
\newcommand{\averageRank}[2]{\cellcolor{RkIndigo#1}#2}

\title{Not Every Cue Is Evidence: Cue Supportability for Forensic Distillation in Face Forgery Detection}

\author{
  \textbf{Binjia Zhou\textsuperscript{1}},
  \textbf{Dawei Luo\textsuperscript{3}},
  \textbf{Shuai Chen\textsuperscript{3}\textsuperscript{\dag}},
  \textbf{Feng Xu\textsuperscript{3}},
  \textbf{Haoyuan Li\textsuperscript{2}},
\\
  \textbf{Jia Wen Seow\textsuperscript{3}},
  \textbf{Jiachi Wang\textsuperscript{1}},
  \textbf{Jie Lei\textsuperscript{4}},
  \textbf{Zunlei Feng\textsuperscript{1}},
  \textbf{Yijun Bei\textsuperscript{1}\textsuperscript{*}}
\\
  \textsuperscript{1}School of Software Technology, Zhejiang University
\\
  \textsuperscript{2}College of Computer Science and Technology, Zhejiang University
\\
  \textsuperscript{3}Ant Group
\\
  \textsuperscript{4}College of Computer Science and Technology, Zhejiang University of Technology
\\
  \small{\textsuperscript{\dag}Project leader. \quad \textsuperscript{*}Corresponding author.}
}

\begin{document}
\maketitle
\begin{abstract}
Explainable face forgery detection requires not only correct judgments but also forensic evidence supported by the input image.
However, training cues are typically derived from the forgery generation pipeline. Since images available to the model are of limited quality and resolution, it is difficult to observe the fine-grained traces these cues refer to. Training methods that further supply such cues as privileged supervision exacerbate the problem, as the student is taught to articulate evidence it cannot perceive. The student consequently resorts to templated or over-generalized expressions, giving rise to the \emph{correct-answer, wrong-evidence} (CAWE) failure.
We propose a dual supportability audit that filters privileged cues before distillation. Each candidate cue must be predictable from the remaining privileged context and causally tied to high-attribution visual regions in the current image. Only cues satisfying both conditions receive substantial distillation weight, preventing unsupported details from leaking into the student.
Experiments across eight benchmarks show that our method maintains or improves detection accuracy while significantly raising explanation faithfulness compared to existing paradigms.
\end{abstract}

\section{Introduction}

\begin{figure*}[t]
  \centering
  \includegraphics[width=\textwidth]{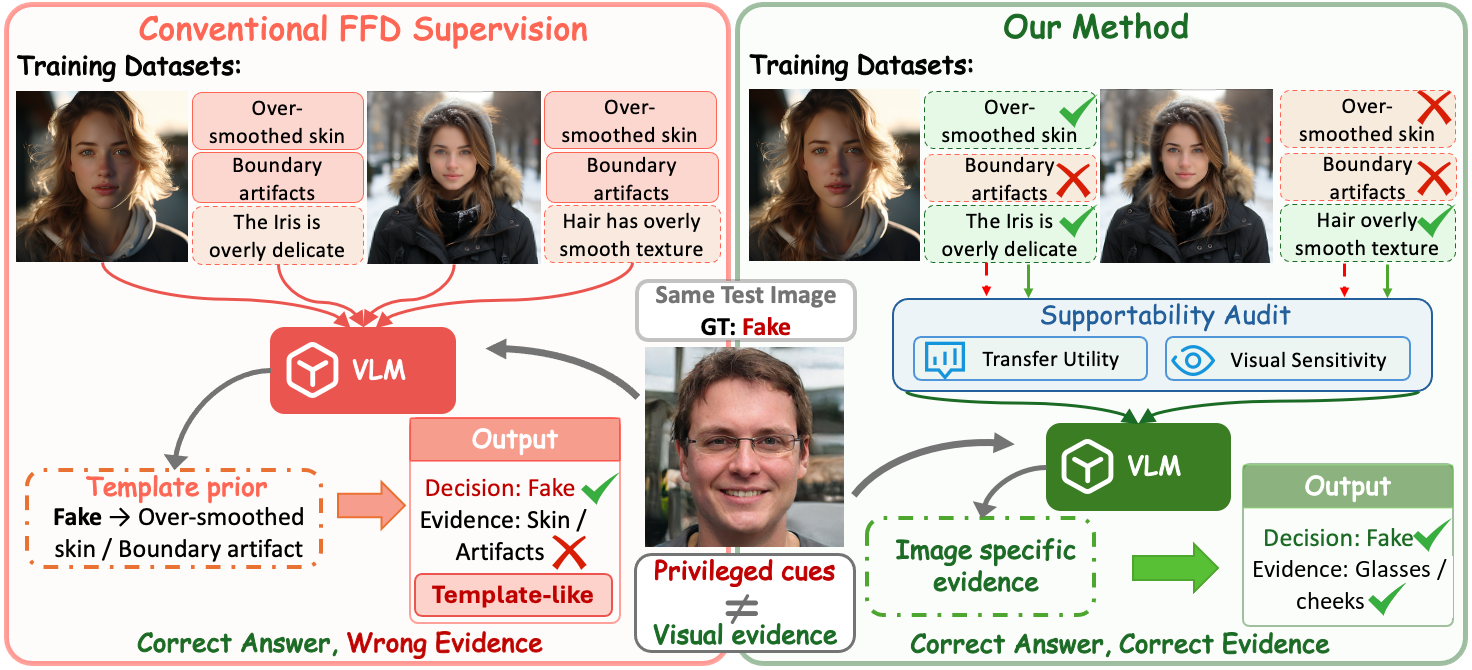}
  \renewcommand{\thefigure}{2}
  \vspace{-1.5em}
  \caption{Comparison between conventional FFD supervision and our method. Conventional approaches train models with privileged cues that cannot be observed from the image, leading to template-like evidence. Our supportability audit filters unsupported cues, enabling image-specific faithful explanations.}
  \vspace{-1.0em}
  \label{fig:teaser}
\end{figure*}

The rapid proliferation of face forgery techniques poses serious threats to privacy, public trust, and information security, making reliable face forgery detection an urgent necessity~\cite{tolosana2020deepfakes,mirsky2021creation}.
Face forgery detection is now moving beyond binary classification~\cite{rossler2019faceforensics++,yan2024df40} toward explainable paradigms that produce natural language evidence~\cite{foteinopoulou2024hitchhiker,huang2024ffaa,sun2025towards,liu2024evolving,wang2024deepfake}. In these settings, a correct real/fake label alone is insufficient; the model must also cite visual anomalies present in the image. Yet a correct answer does not guarantee correct evidence. Recent work~\cite{kuckreja2026pixels} has noted the \emph{correct-answer, wrong-evidence} phenomenon: models predict correctly yet invoke cues unsupported by the current image. Why this failure persists across training paradigms and how to prevent it remain open questions, as illustrated in Fig.~\ref{fig:teaser}.

We argue that the root cause lies in the supervision itself. Existing FFD datasets~\cite{rossler2019faceforensics++,he2021forgerynet,huang2024ffaa,yan2024df40} derive their annotations from the forgery generation pipeline: ground-truth labels, forgery types, manipulation regions, and modification masks all describe how a sample was manufactured. Yet the images available to the model are often of limited quality and resolution, so many manufacturing traces are no longer discernible in the final pixels. Consider a sample annotated as having undergone face swapping in the lower face: the explanation accordingly states ``jawline blending artifact'', yet the actual image may exhibit no visible jawline discontinuity at the given resolution. When such annotations are used as training targets, the model is taught to articulate evidence it cannot perceive, and the supervision reflects how the forgery was made rather than what can be observed.

\begin{figure}[t]
  \centering
  \includegraphics[width=\columnwidth]{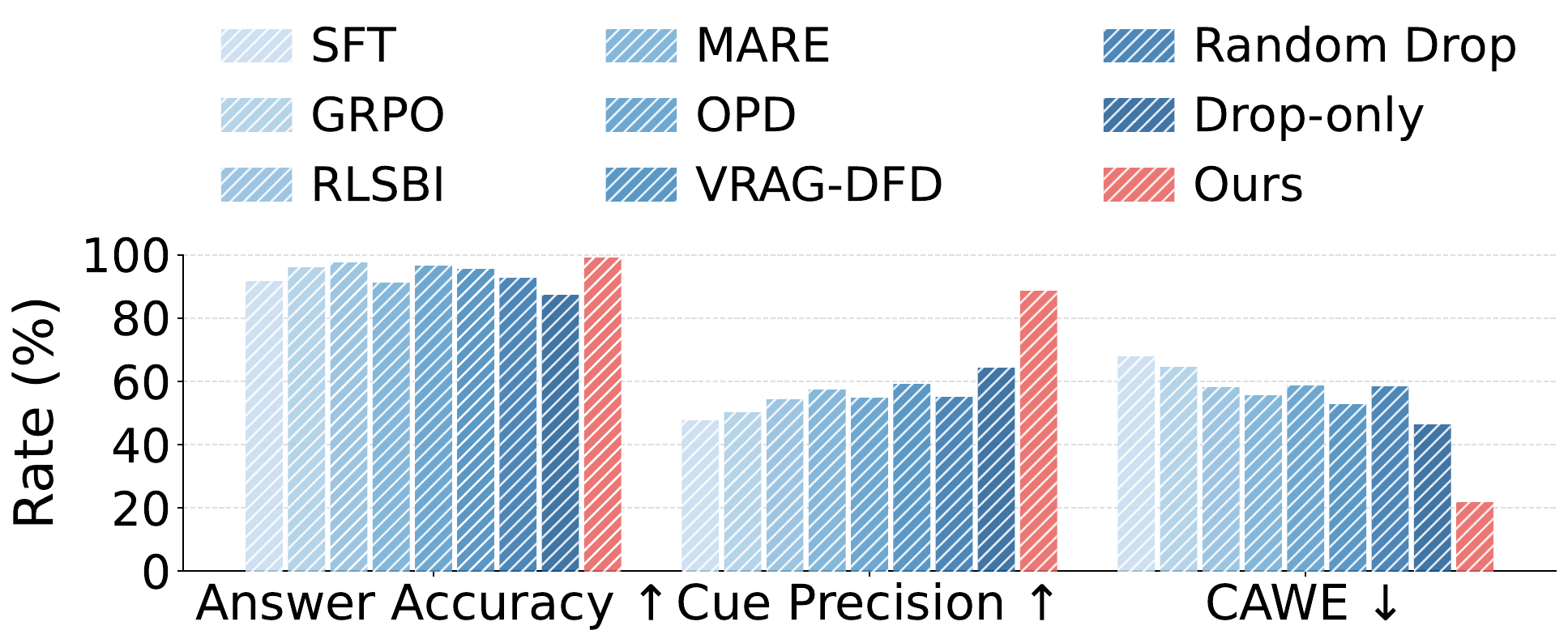}
  \vspace{-1.5em}
  \renewcommand{\thefigure}{1}
  \caption{Comparison of Answer Accuracy, Cue Precision, and CAWE on FF++ across training paradigms and our ablation variants.}
  \label{fig:paradigm_comparison}
  \vspace{-1.5em}
\end{figure}

\begin{figure}[t]
\centering
\includegraphics[width=\columnwidth]{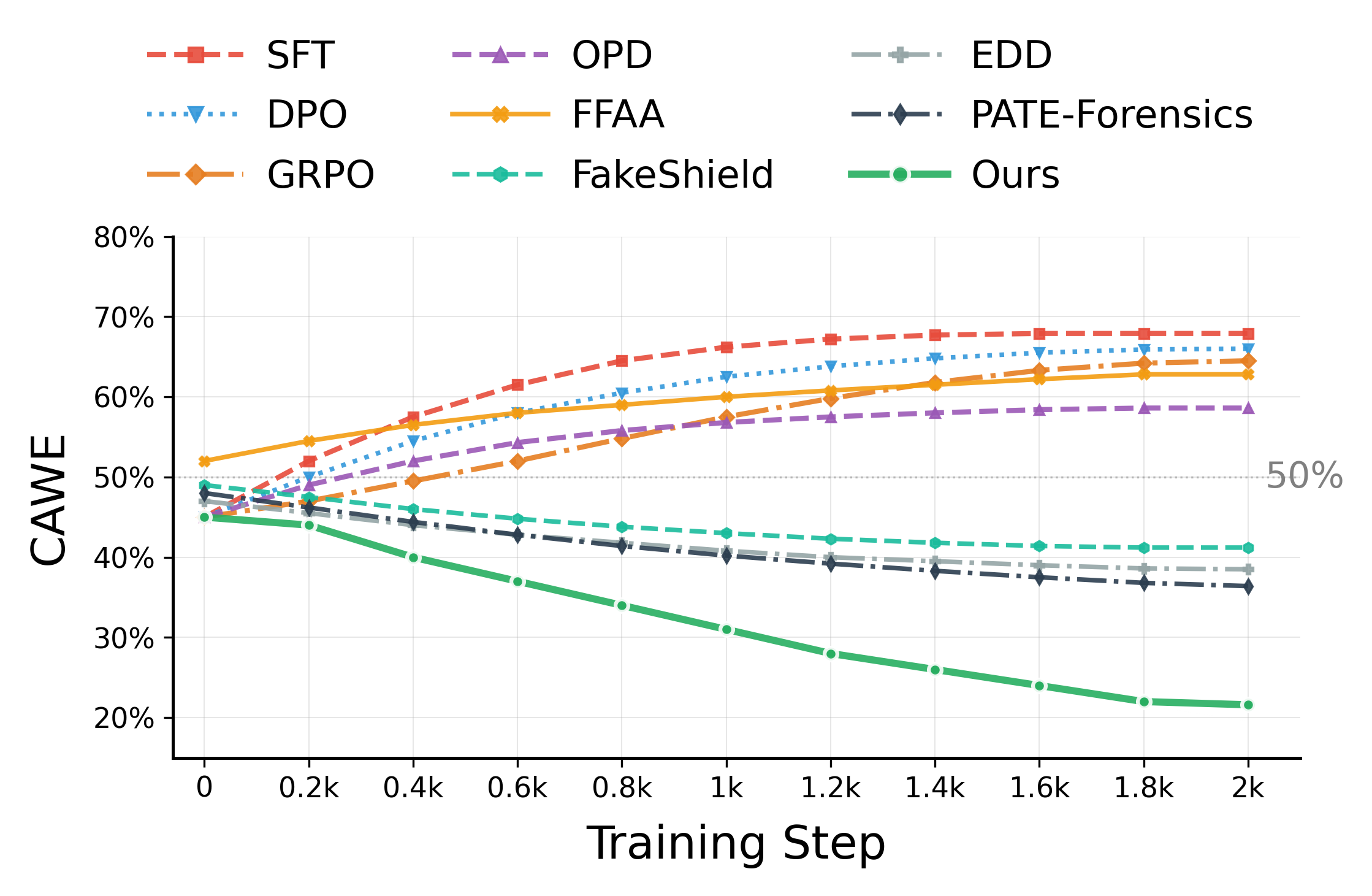}
\vspace{-1.5em}
\caption{CAWE on FF++ across training paradigms; higher means more unfaithful explanations. Most existing methods see CAWE rise with training, while ours consistently reduces it.}
\label{fig:cawe_comparison}
\vspace{-0.5em}
\end{figure}

Several training strategies have been explored to align explanations with visual content. Supervised fine-tuning with label-conditioned rationales~\cite{sun2025towards,huang2024ffaa} degenerates into memorizing label-to-explanation templates, since the label is visible during training. Reinforcement learning~\cite{li2025raidx,huang2025so} may reward explanation format or keyword presence, yet these response-level signals cannot verify whether each cue is supported by the current image. Privileged distillation~\cite{agarwal2024policy} transfers annotation-conditioned knowledge to an image-only student, forcing it to imitate cues unsupported at inference time. Across paradigms, no mechanism checks whether each cue is supportable from the image alone, as shown in Figs.~\ref{fig:paradigm_comparison} and~\ref{fig:cawe_comparison}.

We address this issue by introducing two verification steps before distilling any cue to the student. The first step masks the target cue from the privileged annotation and checks whether the surrounding context alone can still make that cue predictable. A cue that becomes unpredictable once masked was copied verbatim from the annotation and carries no information the student could learn from. The second step perturbs the image regions most relevant to the cue and measures whether the cue likelihood drops accordingly. A cue whose likelihood remains unchanged under such perturbation is not grounded in the current image but instead reflects dataset-level patterns. Only cues that pass both checks receive substantial weight during distillation. This filtering concentrates learning on evidence the student can genuinely perceive from the image alone.
Our contributions are as follows.
\begin{itemize}
\item We trace the correct-answer, wrong-evidence failure to the gap between pipeline annotations and observability, and show that existing training strategies do not close this gap.
\item We propose a dual supportability audit that verifies, for each cue, whether it is predictable from context and sensitive to image content, and distill only cues that pass both checks.
\item Experiments across eight benchmarks show that our method maintains or improves detection accuracy while substantially improving explanation faithfulness.
\end{itemize}

\section{Related Work}

\noindent\textbf{Face Forgery Detection and Explanation.}
Early face forgery detection focuses on binary classification by exploiting spatial artifacts~\cite{li2020face,shiohara2022detecting}, frequency-domain inconsistencies~\cite{qian2020thinking}, or identity-aware cues~\cite{dong2023implicit}. Cross-dataset generalization has been studied through large-scale benchmarks~\cite{rossler2019faceforensics++,he2021forgerynet,yan2024df40,yan2023deepfakebench} that expose distribution shifts across forgery methods.

\begin{figure*}[t]
  \centering
  \includegraphics[width=\textwidth]{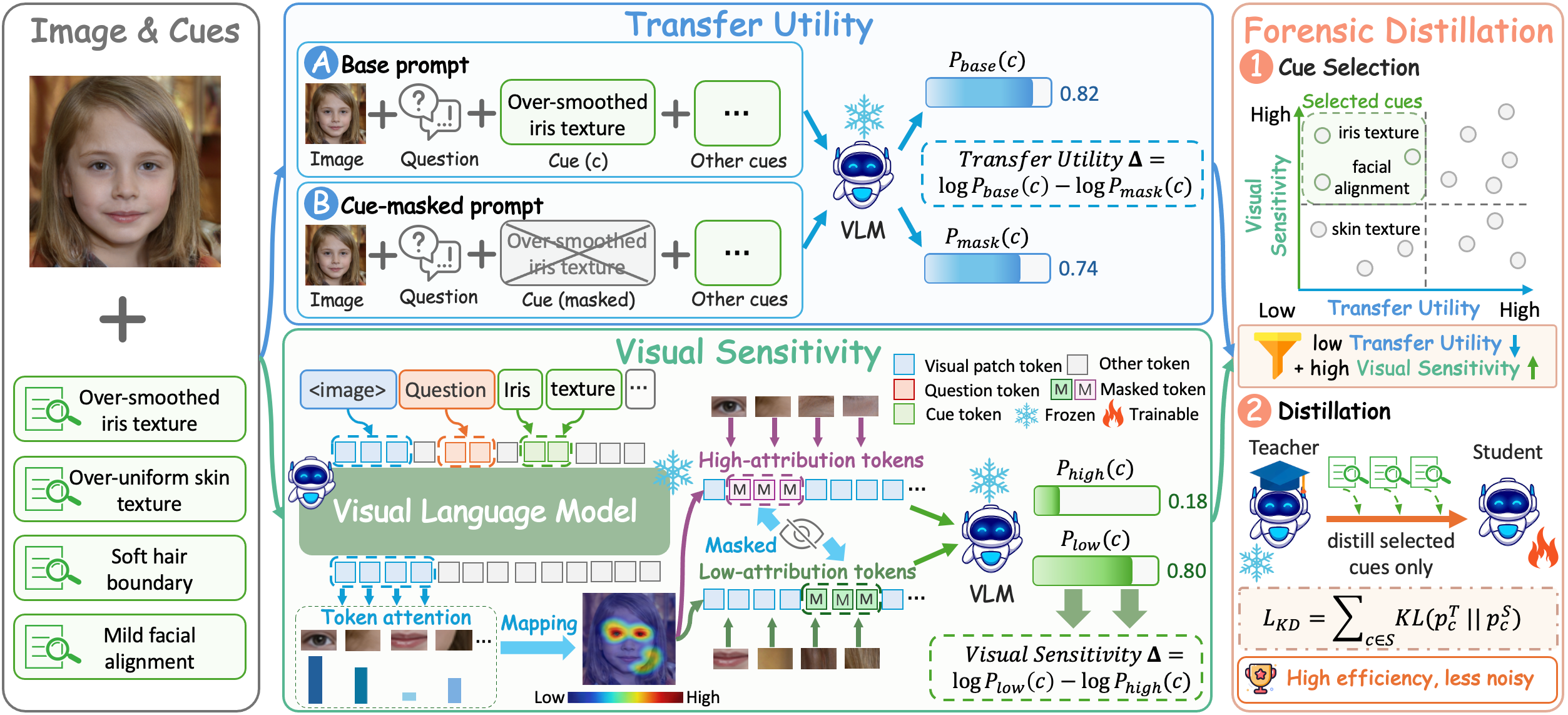}
  \vspace{-1.5em}
  \caption{Overview of our framework. The transfer utility (TU) module (top) masks each candidate cue and measures the likelihood drop, down-weighting cues trivially copied from the annotation. The visual sensitivity (VS) module (bottom) perturbs high-attribution image regions and checks whether cue likelihood is sensitive to visual content. Cues that fail either check receive near-zero weight in the distillation loss.}
  \label{fig:framework}
  \vspace{-0.5em}
\end{figure*}

Advances make detection explainable. Forensic VQA~\cite{foteinopoulou2024hitchhiker}, multimodal report generation~\cite{huang2024ffaa}, and vision-language reasoning~\cite{liu2024evolving,wang2024deepfake} require articulating manipulation evidence; VLFFD~\cite{sun2025towards} pairs detection with visual-linguistic descriptions; Kuckreja et al.~\cite{kuckreja2026pixels} identify the correct-answer, wrong-evidence failure; PATE-Forensics~\cite{li2026pate} leverages perception tools for grounded forensic reasoning. However, explanation supervision derives from manipulation labels, masks, or forgery categories, none of which guarantee instance-level visual observability.

\noindent\textbf{Evidence Faithfulness in Multimodal Models.}
Producing natural-language evidence does not guarantee that it is faithful to visual content. Visual evidence prompting~\cite{praharaj2025reveal,zhuforensic,guo2026grounding} injects structured forensic cues into vision-language models to spatially ground explanations in image regions. RL-based preference methods~\cite{li2025raidx,huang2025so,jiang2026explainable,xu2026mare,han2026vragdfd} apply GRPO or DPO to reward explanations aligned with manipulation traces, improving both detection accuracy and rationale quality. Evidence-assisted generation~\cite{xu2026edd} constructs chosen-rejected explanation pairs and optimizes the model to prefer complete, artifact-grounded evidence over hallucinated or omitted details. These approaches refine or reward evidence after the supervision signal is formed. None examine whether each privileged cue is supportable by the image alone before it enters the training objective.

\noindent\textbf{Post-training and Privileged Distillation.}
Orthogonal to what evidence to trust, the choice of training paradigm shapes explanation quality. Supervised fine-tuning with rationales~\cite{sun2025towards,huang2024ffaa,wei2022chain} trains models to produce chain-of-thought or forensic explanations conditioned on ground-truth labels, but label visibility during training encourages template memorization rather than visual grounding. Policy optimization methods such as PPO~\cite{schulman2017proximal}, DPO~\cite{rafailov2023direct}, and GRPO~\cite{shao2024deepseekmath} instead learn from reward or preference signals. In forensic settings~\cite{li2025raidx,huang2025so}, however, these rewards operate at the response level and do not verify whether cues are visually supported, leaving room for plausible but unfaithful evidence. Privileged knowledge distillation~\cite{hinton2015distilling,vapnik2015learning} lets a teacher with privileged information guide a student with restricted inputs, and on-policy distillation~\cite{agarwal2024policy} trains the student on its own generations under teacher feedback. Concurrent work audits tool outputs~\cite{anonymous2026audit}, verifies teacher fixes by answer recovery~\cite{anonymous2026mvi}, or cancels generic edit effects with matched controls at decoding time~\cite{anonymous2026mcd}, all to improve reasoning correctness. Yet none of these methods test whether each distilled cue is supportable under the student's image-only condition. A dual audit is needed: a context check against cues copied from the annotation and a visual sensitivity check against cues ungrounded in the image.

\section{Method}

Existing explainable FFD methods distill privileged cues indiscriminately, yet many are unsupported because annotations describe how a forgery was manufactured rather than what remains visible. We propose a dual-audit framework that scores each candidate cue for contextual predictability and visual sensitivity before distillation (Fig.~\ref{fig:framework}).

\subsection{Cue Supportability Criterion}
\label{sec:formulation}

\paragraph{Task definition.}
We define the input as an image $I$ paired with a task prompt $q$. The structured annotation is $g=(y,o,r,c)$, where $y$ is the real/fake answer, $o$ is the forgery operation type, $r$ is the manipulation region, and $c$ is the forensic cue. At inference, the student sees only $I$ and $q$; the teacher also sees $o$, $r$, $c$, masks, or source images.

\paragraph{Two failure modes.}
A privileged cue should not be distilled when it exhibits either (i) \emph{direct-cue dependence}: the cue is predictable only when its own text is visible in the input, indicating mere copying; or (ii) \emph{image-independent prior}: the cue likelihood remains unchanged regardless of which image is presented, indicating reliance on annotation frequency rather than visual content.

\paragraph{Unified supportability criterion.}
A cue $c$ is deemed supportable if and only if it exhibits low direct-cue dependence and high visual sensitivity:
\begin{equation*}
\operatorname{Supp}(c) \iff \underbrace{\operatorname{TU}(c) \;\text{is low}}_{\text{low dependence}} \;\land\; \underbrace{\operatorname{VS}(c) \;\text{is high}}_{\text{visual grounding}},
\end{equation*}
where $\operatorname{TU}(c)$ denotes transfer utility and $\operatorname{VS}(c)$ denotes visual sensitivity.
A high TU score indicates the cue is copied from the annotation rather than inferred from context; a low VS score indicates the cue is not grounded in the current image. Both criteria are implemented as soft gates rather than hard thresholds, as detailed in Sec.~\ref{sec:transfer} and \ref{sec:visual}.

\subsection{Transfer Utility}
\label{sec:transfer}

Given the supportability criterion, we operationalize the text-side component. Transfer utility measures how much the teacher relies on directly seeing the cue text to predict it. A high TU indicates strong direct-cue dependence, meaning the cue is largely copied from the annotation rather than inferred from other context.

\paragraph{Teacher distributions.}
For each annotated cue $c$, we teacher-force its tokens after the student-generated prefix. Let $\tilde{y}_{<t}$ denote this combined prefix, where $z_t$ is the annotated token at position $t\in\operatorname{span}(c)$. For the annotation $g=(y,o,r,c)$ defined above, let $g^{-c}=(y,o,r)$ denote the context with the target cue removed. We define a \emph{full-context teacher} that observes $g$:
\begin{equation*}
p_{c}^{\mathrm{base},t} = p_T(\cdot \mid I, q, g, \tilde{y}_{<t}),
\end{equation*}
and a \emph{cue-masked teacher} that observes all annotation fields except the target cue:
\begin{equation*}
p_{c}^{\mathrm{mask},t} = p_T(\cdot \mid I, q, g^{-c}, \tilde{y}_{<t}).
\end{equation*}

\paragraph{Direct-cue dependence.}
For target cue $c$ and token position $t\in\operatorname{span}(c)$, we instantiate $\operatorname{TU}(c)$ at position $t$ as the positive log-likelihood drop caused by removing $c$:
\begin{equation*}
\operatorname{TU}_{c,t} = \left[\log p_{c}^{\mathrm{base},t}(z_t) - \log p_{c}^{\mathrm{mask},t}(z_t)\right]_+.
\end{equation*}
A large $\operatorname{TU}_{c,t}$ indicates the teacher can predict the cue only when the cue text is directly visible, i.e., the cue is being copied rather than inferred from other fields. A zero value means masking the cue causes no prediction loss, suggesting the cue is predictable from the remaining context.

\paragraph{Text-side soft weight.}
Since high TU signals direct copying, we assign lower weight to cues with high dependence:
\begin{equation*}
g_{c,t}^{\mathrm{text}} = \exp\!\left(-\frac{\operatorname{TU}_{c,t}}{\tau_a}\right),
\end{equation*}
where $\tau_a$ is a temperature hyperparameter. This weight approaches one when $\operatorname{TU}_{c,t}=0$, indicating minimal direct-cue dependence, and decays toward zero as dependence increases.

The distillation target uses $p_c^{\mathrm{mask},t}$ rather than the full-context teacher, preventing the literal cue text from leaking into the student. 

\subsection{Visual Sensitivity}
\label{sec:visual}

Low direct-cue dependence confirms that the cue is inferable from the surrounding annotation context, yet it does not guarantee that the cue is grounded in the current image. A cue may be predictable simply because it reflects a dataset-level annotation pattern. To filter such image-independent priors, we introduce a visual sensitivity test that checks whether the cue likelihood is causally tied to the relevant image regions.

\paragraph{Canonical cue likelihood.}
For a given annotation cue $c = (c_1, \ldots, c_{|c|})$, we compute its length-normalized teacher log-likelihood using $I$, $q$, and $g^{-c}$:
\begin{equation*}
\ell_c(I;q,g^{-c}) = \frac{1}{|c|}\sum_{j=1}^{|c|}\log p_T(c_j \mid I,q,g^{-c},c_{<j}).
\end{equation*}

\paragraph{Attribution-guided controlled perturbation.}
Let $V=(v_1,\ldots,v_{|V|})$ denote the visual token sequence produced by the vision encoder. For each visual token $u$, we compute the gradient $\times$ activation attribution:
\begin{equation*}
A_{c,u} = \left\| \frac{\partial \ell_c(I;q,g^{-c})}{\partial v_u} \odot v_u \right\|_1.
\end{equation*}
We then select two equal-size subsets: a high-attribution set $M_c^{\mathrm{high}}$ containing the top-$k$ tokens by $A_{c,u}$, and a low-attribution set $M_c^{\mathrm{low}}$ containing the bottom-$k$ tokens, with $|M_c^{\mathrm{high}}| = |M_c^{\mathrm{low}}| = k$.

Both sets undergo the same perturbation operator $\mathcal{T}$, which zeros out the selected visual tokens:
\begin{equation*}
I^{\mathrm{high}} = \mathcal{T}(I, M_c^{\mathrm{high}}), \quad I^{\mathrm{low}} = \mathcal{T}(I, M_c^{\mathrm{low}}).
\end{equation*}

\paragraph{Differential sensitivity score.}
We instantiate $\operatorname{VS}(c)$ as the difference in cue likelihood between the low-attribution and high-attribution perturbations:
\begin{equation*}
\operatorname{VS}(c) = [\ell_c(I^{\mathrm{low}};q,g^{-c}) - \ell_c(I^{\mathrm{high}};q,g^{-c})]_+.
\end{equation*}
This differential design cancels the generic likelihood drop of any image perturbation. A positive $\operatorname{VS}(c)$ indicates that the cue is sensitive to its attributed regions rather than to arbitrary content.

We map the raw score to a soft visual weight:
\begin{equation*}
g_c^{\mathrm{vis}} = \sigma\!\left(\frac{\operatorname{VS}(c) - m_{\mathrm{pert}}}{T_{\mathrm{pert}}}\right),
\end{equation*}
where $m_{\mathrm{pert}}$ is a margin and $T_{\mathrm{pert}}$ is a temperature. Cues with negligible differential sensitivity receive near-zero visual weight.

\subsection{Selective Distillation}
\label{sec:distill}

With both the text-side weight $g_{c,t}^{\mathrm{text}}$ from Sec.~\ref{sec:transfer} and the visual weight $g_c^{\mathrm{vis}}$ from Sec.~\ref{sec:visual}, we combine them into a joint per-token distillation weight:
\begin{equation*}
w_{c,t} = \mathbf{1}[t \in \operatorname{span}(c)] \cdot g_{c,t}^{\mathrm{text}} \cdot g_c^{\mathrm{vis}}.
\end{equation*}
A cue token receives substantial weight only when it exhibits both low direct-cue dependence and strong visual grounding. The selective distillation loss then takes the form:
\begin{equation*}
\mathcal{L}_{\mathrm{sup}} = \frac{\sum_{c,\,t \in \operatorname{span}(c)} w_{c,t}\, D_{\mathrm{KL}}\!\left(\operatorname{sg}[p_c^{\mathrm{mask},t}] \| p_S^t\right)}{\sum_{c,t} w_{c,t} + \epsilon},
\end{equation*}
where $\operatorname{sg}[\cdot]$ denotes stop-gradient, $p_S^t=p_S(\cdot\mid I,q,\tilde{y}_{<t})$ is the student distribution under the same prefix, and $\epsilon$ is a small constant preventing division by zero. The distillation target is the cue-masked teacher $p_c^{\mathrm{mask},t}$, not the full-context teacher, preventing literal cue text from leaking into the student. The overall training objective combines answer supervision with this selective distillation:
\begin{equation*}
\mathcal{L}_{\mathrm{total}} = \lambda_{\mathrm{ans}}\, \mathrm{CE}(y, p_S^{\mathrm{ans}}) + \lambda_{\mathrm{sup}}\, \mathcal{L}_{\mathrm{sup}}.
\end{equation*}
The cross-entropy term between the ground-truth label $y$ and the student answer distribution $p_S^{\mathrm{ans}}$ supervises only the answer token. Rationale tokens receive no unweighted self-imitation signal; their learning is governed entirely by the supportability-weighted KL term. Both the teacher distribution and the audit weights are stop-gradiented with respect to the student. 

\begin{table*}[t]
\centering
\renewcommand{\arraystretch}{1.1}
\resizebox{\textwidth}{!}{%
\begin{tabular}{ll cccc cccc cccc cccc}
\toprule
& & \multicolumn{4}{c}{\textbf{FF++}} & \multicolumn{4}{c}{\textbf{Celeb-DF v2}} & \multicolumn{4}{c}{\textbf{DFDC}} & \multicolumn{4}{c}{\textbf{WildDF}} \\
\cmidrule(lr){3-6} \cmidrule(lr){7-10} \cmidrule(lr){11-14} \cmidrule(lr){15-18}
\textbf{Category} & \textbf{Method} & ACC & AUC & EER$\downarrow$ & CAWE$\downarrow$ & ACC & AUC & EER$\downarrow$ & CAWE$\downarrow$ & ACC & AUC & EER$\downarrow$ & CAWE$\downarrow$ & ACC & AUC & EER$\downarrow$ & CAWE$\downarrow$ \\
\midrule
\multirow{3}{*}{Traditional}
& Xception & \ffRank{A}{95.73} & \ffRank{A}{96.30} & \ffRank{A}{9.52} & -- & \celebRank{A}{90.34} & \celebRank{A}{89.75} & \celebRank{A}{15.20} & -- & \dfdcRank{A}{79.32} & \dfdcRank{A}{81.58} & \dfdcRank{A}{24.50} & -- & \wildRank{A}{75.26} & \wildRank{A}{80.89} & \wildRank{A}{26.80} & -- \\
& SBI & \ffRank{A}{97.80} & \ffRank{B}{99.64} & \ffRank{A}{2.40} & -- & \celebRank{A}{92.45} & \celebRank{A}{93.18} & \celebRank{A}{11.60} & -- & \dfdcRank{A}{82.50} & \dfdcRank{A}{87.30} & \dfdcRank{A}{19.80} & -- & \wildRank{A}{78.40} & \wildRank{A}{84.50} & \wildRank{A}{22.50} & -- \\
& Forensics Adapter & \ffRank{A}{98.35} & \ffRank{C}{99.71} & \ffRank{C}{1.82} & -- & \celebRank{A}{94.20} & \celebRank{A}{95.80} & \celebRank{A}{9.40} & -- & \dfdcRank{A}{85.10} & \dfdcRank{A}{90.20} & \dfdcRank{A}{16.50} & -- & \wildRank{A}{80.60} & \wildRank{A}{87.30} & \wildRank{A}{19.80} & -- \\
\midrule
\multirow{10}{*}{Explainable}
& FFAA & \ffRank{A}{94.80} & \ffRank{A}{96.50} & \ffRank{A}{8.75} & \ffRank{A}{62.8} & \celebRank{A}{88.50} & \celebRank{A}{90.20} & \celebRank{A}{16.80} & \celebRank{A}{68.4} & \dfdcRank{A}{76.30} & \dfdcRank{A}{80.50} & \dfdcRank{A}{26.20} & \dfdcRank{A}{65.2} & \wildRank{A}{72.40} & \wildRank{A}{78.60} & \wildRank{A}{28.50} & \wildRank{A}{71.6} \\
& VLFFD & \ffRank{A}{97.20} & \ffRank{A}{99.23} & \ffRank{A}{3.85} & \ffRank{A}{56.3} & \celebRank{A}{93.80} & \celebRank{A}{95.40} & \celebRank{A}{10.20} & \celebRank{A}{60.1} & \dfdcRank{A}{83.60} & \dfdcRank{A}{88.50} & \dfdcRank{A}{18.40} & \dfdcRank{A}{62.4} & \wildRank{A}{79.50} & \wildRank{A}{85.80} & \wildRank{A}{21.30} & \wildRank{A}{65.8} \\
& Identity-Aware VLM & \ffRank{A}{96.50} & \ffRank{A}{98.40} & \ffRank{A}{5.20} & \ffRank{A}{53.6} & \celebRank{A}{94.25} & \celebRank{A}{96.10} & \celebRank{A}{8.80} & \celebRank{A}{55.4} & \dfdcRank{A}{84.20} & \dfdcRank{A}{89.60} & \dfdcRank{A}{17.20} & \dfdcRank{A}{61.7} & \wildRank{A}{80.30} & \wildRank{A}{86.50} & \wildRank{A}{20.50} & \wildRank{A}{59.2} \\
& RLSBI & \ffRank{A}{97.60} & \ffRank{A}{99.50} & \ffRank{A}{3.10} & \ffRank{A}{58.1} & \celebRank{A}{95.10} & \celebRank{A}{96.80} & \celebRank{A}{7.60} & \celebRank{A}{63.8} & \dfdcRank{A}{85.80} & \dfdcRank{A}{91.40} & \dfdcRank{A}{15.20} & \dfdcRank{A}{62.0} & \wildRank{A}{81.50} & \wildRank{A}{88.20} & \wildRank{A}{18.60} & \wildRank{A}{68.4} \\
& SIDA & \ffRank{A}{97.82} & \ffRank{A}{98.92} & \ffRank{A}{4.80} & \ffRank{A}{54.5} & \celebRank{A}{94.60} & \celebRank{A}{96.50} & \celebRank{A}{8.20} & \celebRank{A}{58.6} & \dfdcRank{A}{85.40} & \dfdcRank{A}{91.20} & \dfdcRank{A}{15.60} & \dfdcRank{A}{63.8} & \wildRank{A}{81.20} & \wildRank{A}{87.80} & \wildRank{A}{19.40} & \wildRank{A}{61.0} \\
& FakeVLM & \ffRank{A}{98.20} & \ffRank{A}{99.10} & \ffRank{A}{3.65} & \ffRank{A}{48.7} & \celebRank{A}{96.80} & \celebRank{A}{98.20} & \celebRank{A}{5.80} & \celebRank{A}{51.2} & \dfdcRank{A}{90.40} & \dfdcRank{A}{95.80} & \dfdcRank{A}{8.60} & \dfdcRank{A}{54.8} & \wildRank{A}{84.30} & \wildRank{A}{91.50} & \wildRank{A}{13.80} & \wildRank{A}{52.4} \\
& M2F2-Det & \ffRank{A}{98.79} & \ffRank{A}{99.34} & \ffRank{A}{2.95} & \ffRank{A}{50.4} & \celebRank{B}{98.98} & \celebRank{D}{99.92} & \celebRank{A}{2.15} & \celebRank{A}{52.1} & \dfdcRank{A}{92.80} & \dfdcRank{A}{97.50} & \dfdcRank{A}{6.20} & \dfdcRank{A}{58.6} & \wildRank{A}{86.05} & \wildRank{A}{93.14} & \wildRank{A}{11.50} & \wildRank{A}{55.3} \\
& FakeShield & \ffRank{A}{98.35} & \ffRank{A}{99.15} & \ffRank{A}{3.60} & \ffRank{A}{41.2} & \celebRank{A}{97.50} & \celebRank{A}{98.80} & \celebRank{A}{4.50} & \celebRank{A}{43.8} & \dfdcRank{A}{92.10} & \dfdcRank{A}{97.20} & \dfdcRank{A}{6.80} & \dfdcRank{A}{49.5} & \wildRank{A}{85.80} & \wildRank{A}{92.60} & \wildRank{A}{12.40} & \wildRank{A}{46.7} \\
& EDD & \ffRank{C}{99.05} & \ffRank{A}{99.52} & \ffRank{A}{2.10} & \ffRank{B}{38.5} & \celebRank{C}{99.10} & \celebRank{A}{99.78} & \celebRank{B}{2.05} & \celebRank{B}{40.2} & \dfdcRank{C}{93.90} & \dfdcRank{C}{98.20} & \dfdcRank{C}{5.40} & \dfdcRank{B}{42.6} & \wildRank{C}{87.80} & \wildRank{C}{94.20} & \wildRank{C}{10.50} & \wildRank{B}{39.8} \\
& PATE-Forensics & \ffRank{B}{98.90} & \ffRank{A}{99.60} & \ffRank{B}{1.95} & \ffRank{C}{36.4} & \celebRank{A}{98.85} & \celebRank{B}{99.85} & \celebRank{C}{1.90} & \celebRank{C}{35.6} & \dfdcRank{B}{93.50} & \dfdcRank{B}{98.10} & \dfdcRank{B}{5.60} & \dfdcRank{C}{38.2} & \wildRank{B}{87.20} & \wildRank{B}{93.80} & \wildRank{B}{10.80} & \wildRank{C}{36.5} \\
\midrule
& \textbf{Ours} & \ffRank{D}{99.12} & \ffRank{D}{99.78} & \ffRank{D}{1.65} & \ffRank{D}{21.6} & \celebRank{D}{99.35} & \celebRank{C}{99.86} & \celebRank{D}{1.72} & \celebRank{D}{22.8} & \dfdcRank{D}{94.60} & \dfdcRank{D}{98.45} & \dfdcRank{D}{5.10} & \dfdcRank{D}{27.4} & \wildRank{D}{88.50} & \wildRank{D}{94.82} & \wildRank{D}{9.85} & \wildRank{D}{24.1} \\
& \textit{Gain} & $+0.07$ & $+0.07$ & $+0.17$ & $+14.8$ & $+0.25$ & $-0.06$ & $+0.18$ & $+12.8$ & $+0.70$ & $+0.25$ & $+0.30$ & $+10.8$ & $+0.70$ & $+0.62$ & $+0.65$ & $+12.4$ \\
\bottomrule
\end{tabular}%
}
\vspace{-0.5em}
\caption{Main results on four benchmarks. Color intensity indicates SOTA ranking. The Gain row reports improvements over the strongest baseline (reductions for EER and CAWE); a negative value indicates underperformance. CAWE is reported only for methods that generate forensic evidence. Our method achieves competitive detection accuracy while substantially reducing CAWE over all explainable baselines.}
\label{tab:main}
\vspace{-0.5em}
\end{table*}

\begin{table*}[t]
\centering
\renewcommand{\arraystretch}{1.1}
\resizebox{\textwidth}{!}{%
\begin{tabular}{l cccc cccc cccc cccc}
\toprule
& \multicolumn{4}{c}{\textbf{DeeperForensics}} & \multicolumn{4}{c}{\textbf{Celeb-DF v2}} & \multicolumn{4}{c}{\textbf{ForgeryNet}} & \multicolumn{4}{c}{\textbf{Celeb-DF++}} \\
\cmidrule(lr){2-5} \cmidrule(lr){6-9} \cmidrule(lr){10-13} \cmidrule(lr){14-17}
\textbf{Method} & Tok.$\downarrow$ & Inf.$\downarrow$ & ACC & CAWE$\downarrow$ & Tok.$\downarrow$ & Inf.$\downarrow$ & ACC & CAWE$\downarrow$ & Tok.$\downarrow$ & Inf.$\downarrow$ & ACC & CAWE$\downarrow$ & Tok.$\downarrow$ & Inf.$\downarrow$ & ACC & CAWE$\downarrow$ \\
\midrule
Standard SFT & \deeperRank{D}{6.8M} & \deeperRank{A}{842} & \deeperRank{A}{89.45} & \deeperRank{A}{74.2} & \celebRank{D}{4.2M} & \celebRank{A}{856} & \celebRank{A}{91.20} & \celebRank{A}{71.6} & \forgeryRank{D}{12.4M} & \forgeryRank{A}{810} & \forgeryRank{A}{74.80} & \forgeryRank{A}{79.8} & \celebPlusRank{D}{5.6M} & \celebPlusRank{A}{838} & \celebPlusRank{A}{90.65} & \celebPlusRank{A}{73.5} \\
Self-Correction SFT & \deeperRank{A}{15.2M} & \deeperRank{A}{1,104} & \deeperRank{A}{90.80} & \deeperRank{A}{76.5} & \celebRank{A}{9.4M} & \celebRank{A}{1,126} & \celebRank{A}{92.60} & \celebRank{A}{73.4} & \forgeryRank{A}{27.8M} & \forgeryRank{A}{1,068} & \forgeryRank{A}{76.20} & \forgeryRank{A}{81.4} & \celebPlusRank{A}{12.5M} & \celebPlusRank{A}{1,092} & \celebPlusRank{A}{91.85} & \celebPlusRank{A}{75.8} \\
LLaVA-CoT & \deeperRank{A}{32.1M} & \deeperRank{A}{1,158} & \deeperRank{A}{92.30} & \deeperRank{A}{66.8} & \celebRank{A}{19.8M} & \celebRank{A}{1,182} & \celebRank{A}{94.80} & \celebRank{A}{64.2} & \forgeryRank{A}{58.6M} & \forgeryRank{A}{1,120} & \forgeryRank{A}{80.40} & \forgeryRank{A}{74.5} & \celebPlusRank{A}{26.4M} & \celebPlusRank{A}{1,146} & \celebPlusRank{A}{94.10} & \celebPlusRank{A}{67.6} \\
Standard GRPO & \deeperRank{A}{89.4M} & \deeperRank{A}{1,318} & \deeperRank{A}{93.80} & \deeperRank{A}{70.4} & \celebRank{A}{55.2M} & \celebRank{A}{1,346} & \celebRank{A}{95.60} & \celebRank{A}{68.2} & \forgeryRank{A}{163.2M} & \forgeryRank{A}{1,274} & \forgeryRank{A}{82.40} & \forgeryRank{A}{75.8} & \celebPlusRank{A}{73.6M} & \celebPlusRank{A}{1,302} & \celebPlusRank{A}{95.20} & \celebPlusRank{A}{72.6} \\
DAPO & \deeperRank{A}{114.7M} & \deeperRank{A}{1,280} & \deeperRank{A}{94.10} & \deeperRank{A}{71.8} & \celebRank{A}{70.8M} & \celebRank{A}{1,304} & \celebRank{A}{95.30} & \celebRank{A}{69.4} & \forgeryRank{A}{209.4M} & \forgeryRank{A}{1,238} & \forgeryRank{A}{81.90} & \forgeryRank{A}{78.2} & \celebPlusRank{A}{94.4M} & \celebPlusRank{A}{1,266} & \celebPlusRank{A}{94.80} & \celebPlusRank{A}{73.0} \\
Dr.GRPO & \deeperRank{A}{82.6M} & \deeperRank{A}{1,286} & \deeperRank{A}{94.20} & \deeperRank{A}{68.5} & \celebRank{A}{51.0M} & \celebRank{A}{1,310} & \celebRank{A}{95.80} & \celebRank{A}{66.8} & \forgeryRank{A}{150.8M} & \forgeryRank{A}{1,242} & \forgeryRank{A}{83.10} & \forgeryRank{A}{73.6} & \celebPlusRank{A}{68.0M} & \celebPlusRank{A}{1,268} & \celebPlusRank{A}{95.50} & \celebPlusRank{A}{70.2} \\
GSPO & \deeperRank{A}{78.2M} & \deeperRank{A}{1,248} & \deeperRank{A}{94.65} & \deeperRank{A}{67.2} & \celebRank{A}{48.3M} & \celebRank{A}{1,272} & \celebRank{A}{96.10} & \celebRank{A}{65.4} & \forgeryRank{A}{142.8M} & \forgeryRank{A}{1,206} & \forgeryRank{A}{83.80} & \forgeryRank{A}{72.0} & \celebPlusRank{A}{64.4M} & \celebPlusRank{A}{1,230} & \celebPlusRank{A}{95.85} & \celebPlusRank{A}{68.8} \\
SRPO & \deeperRank{A}{52.4M} & \deeperRank{A}{1,186} & \deeperRank{A}{94.85} & \deeperRank{A}{66.0} & \celebRank{A}{32.4M} & \celebRank{A}{1,208} & \celebRank{A}{96.50} & \celebRank{A}{64.2} & \forgeryRank{A}{95.8M} & \forgeryRank{A}{1,148} & \forgeryRank{A}{84.60} & \forgeryRank{A}{71.8} & \celebPlusRank{A}{43.2M} & \celebPlusRank{A}{1,172} & \celebPlusRank{A}{96.20} & \celebPlusRank{A}{67.4} \\
OPD & \deeperRank{A}{12.4M} & \deeperRank{A}{876} & \deeperRank{A}{94.50} & \deeperRank{A}{63.5} & \celebRank{A}{7.6M} & \celebRank{A}{892} & \celebRank{A}{96.20} & \celebRank{A}{61.8} & \forgeryRank{A}{22.6M} & \forgeryRank{A}{848} & \forgeryRank{A}{83.60} & \forgeryRank{A}{72.4} & \celebPlusRank{A}{10.2M} & \celebPlusRank{A}{864} & \celebPlusRank{A}{95.80} & \celebPlusRank{A}{60.6} \\
Lightning OPD & \deeperRank{A}{10.8M} & \deeperRank{A}{834} & \deeperRank{A}{95.10} & \deeperRank{A}{62.8} & \celebRank{A}{6.6M} & \celebRank{A}{850} & \celebRank{A}{96.85} & \celebRank{A}{60.4} & \forgeryRank{A}{19.7M} & \forgeryRank{A}{806} & \forgeryRank{A}{85.20} & \forgeryRank{A}{71.6} & \celebPlusRank{A}{8.9M} & \celebPlusRank{A}{822} & \celebPlusRank{A}{96.30} & \celebPlusRank{A}{63.2} \\
OPRD & \deeperRank{A}{11.4M} & \deeperRank{B}{802} & \deeperRank{A}{95.60} & \deeperRank{A}{61.5} & \celebRank{A}{7.0M} & \celebRank{B}{818} & \celebRank{A}{97.20} & \celebRank{A}{59.8} & \forgeryRank{A}{20.8M} & \forgeryRank{B}{774} & \forgeryRank{A}{86.40} & \forgeryRank{A}{70.2} & \celebPlusRank{A}{9.4M} & \celebPlusRank{B}{790} & \celebPlusRank{A}{96.70} & \celebPlusRank{A}{60.4} \\
G-OPD & \deeperRank{A}{13.2M} & \deeperRank{A}{862} & \deeperRank{B}{96.10} & \deeperRank{B}{60.8} & \celebRank{A}{8.2M} & \celebRank{A}{878} & \celebRank{B}{97.60} & \celebRank{B}{58.6} & \forgeryRank{A}{24.2M} & \forgeryRank{A}{836} & \forgeryRank{B}{87.20} & \forgeryRank{B}{69.4} & \celebPlusRank{A}{10.8M} & \celebPlusRank{A}{850} & \celebPlusRank{B}{97.10} & \celebPlusRank{B}{59.0} \\
SDPO & \deeperRank{B}{10.2M} & \deeperRank{C}{788} & \deeperRank{C}{96.30} & \deeperRank{C}{58.6} & \celebRank{B}{6.3M} & \celebRank{C}{804} & \celebRank{C}{97.85} & \celebRank{C}{56.2} & \forgeryRank{B}{18.6M} & \forgeryRank{C}{762} & \forgeryRank{C}{87.80} & \forgeryRank{C}{66.8} & \celebPlusRank{B}{8.4M} & \celebPlusRank{C}{776} & \celebPlusRank{C}{97.40} & \celebPlusRank{C}{57.0} \\
\midrule
\textbf{Ours} & \deeperRank{C}{9.4M} & \deeperRank{D}{547} & \deeperRank{D}{97.25} & \deeperRank{D}{28.6} & \celebRank{C}{5.8M} & \celebRank{D}{562} & \celebRank{D}{99.10} & \celebRank{D}{24.2} & \forgeryRank{C}{17.2M} & \forgeryRank{D}{528} & \forgeryRank{D}{90.80} & \forgeryRank{D}{38.4} & \celebPlusRank{C}{7.7M} & \celebPlusRank{D}{540} & \celebPlusRank{D}{98.85} & \celebPlusRank{D}{30.8} \\
\bottomrule
\end{tabular}%
}
\vspace{-0.5em}
\caption{Comparison across paradigms on four benchmarks. Tok.\ denotes student-side training tokens (rollouts and student forward passes) drawn from each benchmark's training split; teacher and audit passes, which roughly double our per-step cost, are excluded. Inf.\ denotes average inference tokens per sample. Ours achieves the highest accuracy and lowest CAWE across datasets with fewer student-side tokens than all other distillation and RL alternatives and shorter responses. Shading follows Table~\ref{tab:main}; fewer tokens are better.}
\label{tab:paradigm}
\vspace{-1.0em}
\end{table*}

\section{Experiments}
\label{sec:experiments}


\paragraph{Datasets.}
Across the main, paradigm, and extended comparisons, we evaluate on eight unique face forgery detection benchmarks spanning diverse forgery types and difficulty levels.
For the main comparison (Table~\ref{tab:main}), we use four widely adopted datasets:
FaceForensics++~\cite{rossler2019faceforensics++}, which contains 1,000 original videos manipulated by four methods;
Celeb-DF v2~\cite{li2020celeb}, featuring high-quality celebrity face swaps;
DFDC~\cite{dolhansky2020deepfake}, a large-scale challenge dataset with diverse perturbations; and
WildDF~\cite{zi2020wilddeepfake}, which collects in-the-wild forgeries from online sources.
The training paradigm comparison (Table~\ref{tab:paradigm}) reuses Celeb-DF v2 and additionally includes
DeeperForensics~\cite{jiang2020deeperforensics},
ForgeryNet~\cite{he2021forgerynet}, and
Celeb-DF++~\cite{li2507celeb};
DF40~\cite{yan2024df40} is evaluated in Appendix~\ref{sec:extended_comparison}.

\paragraph{Evaluation metrics.}
We report four metrics.
ACC and AUC measure detection performance.
EER, the Equal Error Rate, captures the operating point where false acceptance equals false rejection.
CAWE, the Correct-Answer Wrong-Evidence rate, quantifies the proportion of correctly classified samples whose forensic evidence is unsupported by the input image.
For the paradigm comparison, we additionally report Tok., the student-side training tokens, and Inf., the average inference tokens per sample, to assess computational efficiency. 

\begin{figure*}[t]
  \centering
  \includegraphics[width=\textwidth]{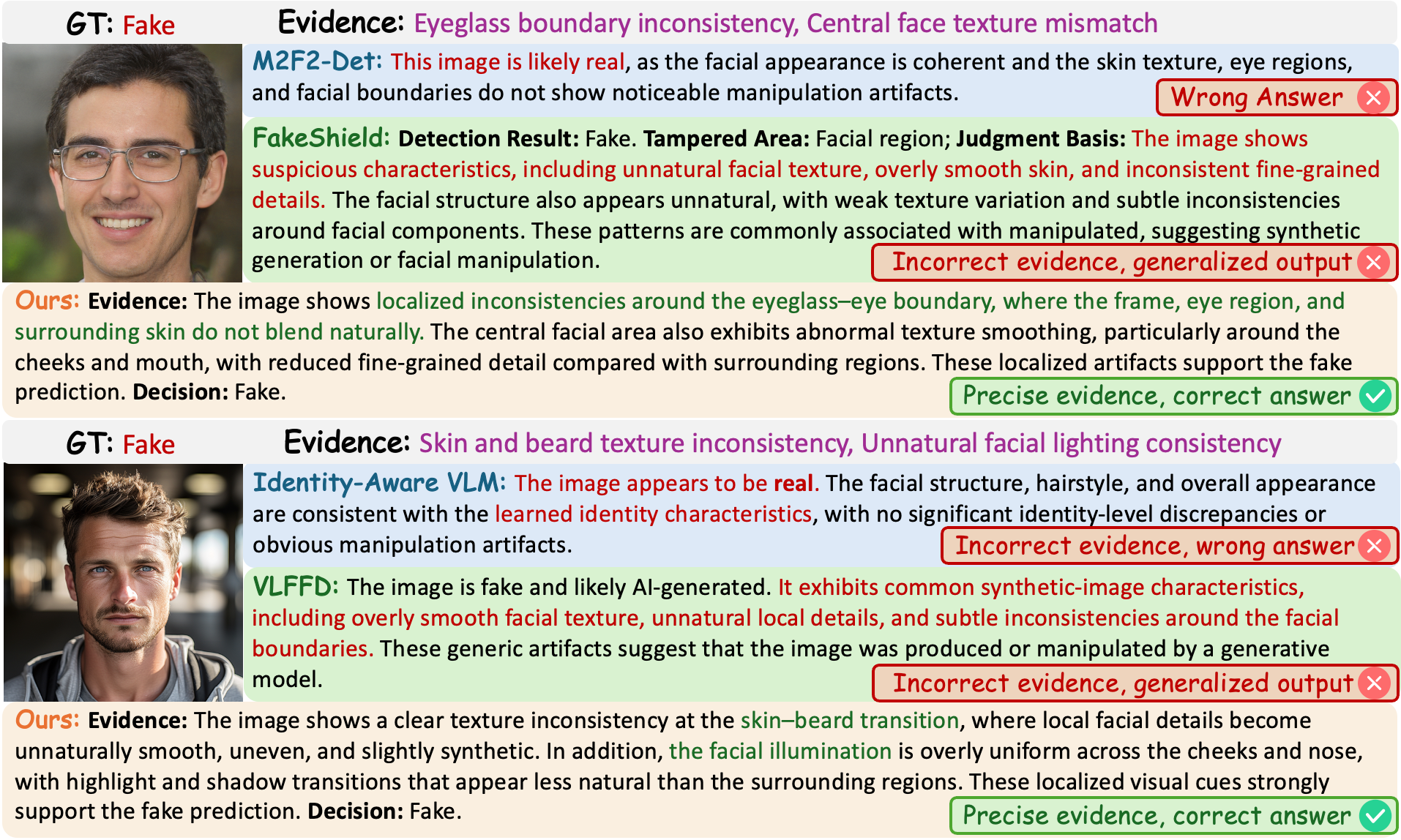}
  \vspace{-1.5em}
  \caption{Qualitative comparison on two forged samples; purple text lists the annotated artifacts. Existing methods either predict the wrong answer or produce template-like evidence unrelated to the specific image. Our method generates precise, image-specific forensic cues that are visually grounded and support the decision.}
  \label{fig:vis}
  \vspace{-1.0em}
\end{figure*}

\paragraph{Baselines.}
We compare against three categories of methods.
(1) \emph{Traditional detectors}: Xception~\cite{chollet2017xception}, SBI~\cite{shiohara2022detecting}, and Forensics Adapter~\cite{cui2025forensics}, which perform binary classification without generating explanations.
(2) \emph{Explainable detectors}: FFAA~\cite{huang2024ffaa}, VLFFD~\cite{sun2025towards}, Identity-Aware VLM~\cite{xu2026identity}, RLSBI~\cite{jiang2026explainable}, SIDA~\cite{huang2025sida}, FakeVLM~\cite{wen2026spot}, M2F2-Det~\cite{guo2025rethinking}, FakeShield~\cite{xu2025fakeshield}, EDD~\cite{xu2026edd}, and PATE-Forensics~\cite{li2026pate}, which produce natural language forensic evidence alongside detection decisions.
(3) \emph{Training paradigms} (Table~\ref{tab:paradigm}): Standard SFT, Self-Correction SFT~\cite{yang2025confidence}, LLaVA-CoT~\cite{xu2025llava}, Standard GRPO~\cite{shao2024deepseekmath}, DAPO~\cite{yu2026dapo}, Dr.GRPO~\cite{liu2025understanding}, GSPO~\cite{chujie2025group}, SRPO~\cite{liu2026srpo}, OPD~\cite{agarwal2024policy}, Lightning OPD~\cite{wu2026lightning}, OPRD~\cite{kim2026coverage}, G-OPD~\cite{yang2026learning}, and SDPO~\cite{kong2025sdpo}.

\paragraph{Implementation details.}
We adopt Qwen3-VL-8B~\cite{bai2025qwen3} as the student backbone. The frozen teacher shares the same architecture and receives the full structured annotation as additional context. We set the TU temperature $\tau_a = 0.5$, the visual sensitivity margin $m_{\mathrm{pert}} = 3.0$, temperature $T_{\mathrm{pert}} = 0.3$, and the attribution subset size $k = 32$ visual tokens. The loss weights are $\lambda_{\mathrm{ans}} = 1.0$ and $\lambda_{\mathrm{sup}} = 0.5$. All models are trained for 3 epochs with a learning rate of $2 \times 10^{-5}$ and a batch size of 32 on 16 NVIDIA H20 GPUs. 

\begin{figure}[t]
  \centering
  \includegraphics[width=\columnwidth]{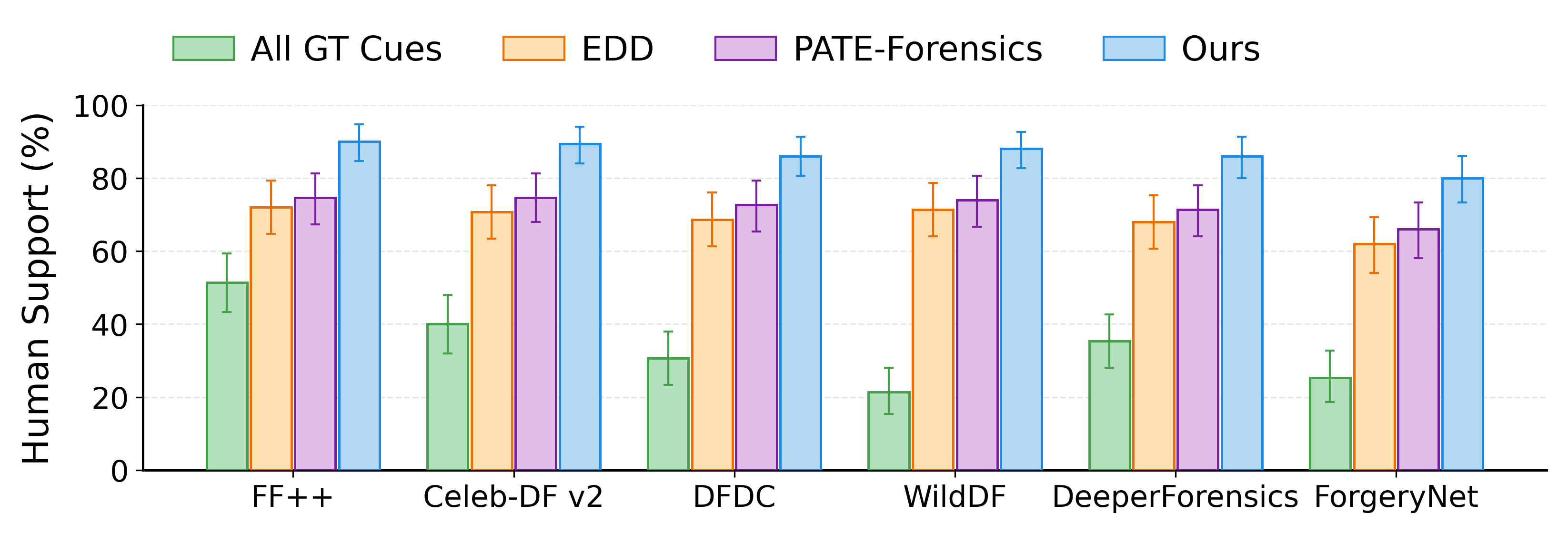}
  \vspace{-1.5em}
  \caption{Human-judged support ratio (\%) of forensic cues on six benchmarks (150 image-cue pairs per cue source and benchmark). Error bars: 95\% bootstrap CIs.}
  \label{fig:human_observability}
  \vspace{-1.0em}
\end{figure}

\subsection{Quantitative Results}

\paragraph{Main comparison.}
As shown in Table~\ref{tab:main}, our method achieves the best or second-best detection performance across all four benchmarks in ACC, AUC, and EER, on par with the strongest explainable baselines EDD and PATE-Forensics. The key advantage lies in explanation faithfulness: our CAWE is reduced by 28 to 41\% relative to the strongest competing method across all four benchmarks, showing that the audit removes visually unsupported cues without sacrificing accuracy. Traditional detectors reach competitive AUC but cannot produce forensic evidence.

\paragraph{Training paradigm comparison.}
Table~\ref{tab:paradigm} compares our method against 13 training paradigms across four benchmarks. Our method achieves the highest ACC on every dataset while attaining the lowest CAWE. RL methods such as GRPO, DAPO, and Dr.GRPO improve accuracy over SFT but consume 9 to 12$\times$ more training tokens than ours and produce substantially longer responses, yet their CAWE remains high, ranging from 67 to 78\%. Our method achieves the strongest trade-off between accuracy and faithfulness with fewer student-side tokens than all other distillation and RL alternatives, and generates the shortest inference responses, as down-weighting unsupported cues during training reduces computational cost and output verbosity.

\begin{figure*}[t]
  \centering
  \includegraphics[width=\textwidth]{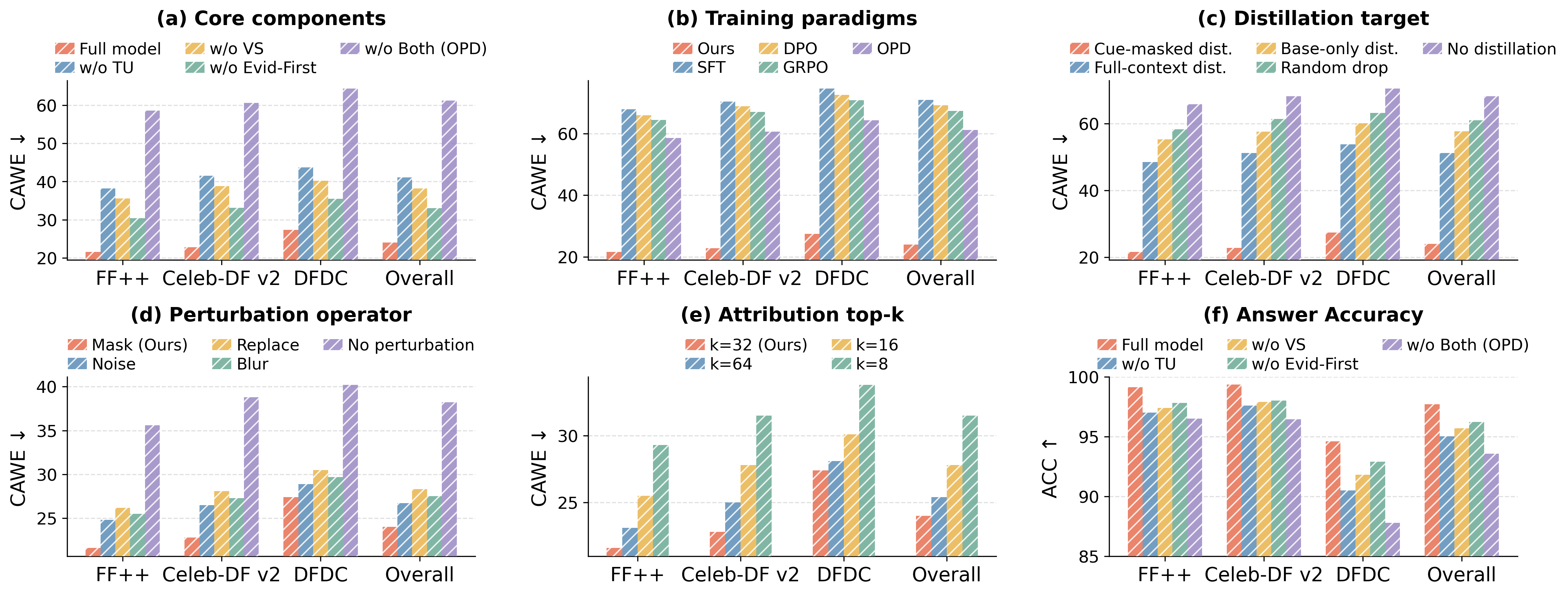}
  \vspace{-1.5em}
  \caption{Ablation results on FF++, Celeb-DF v2, DFDC, and overall average. (a) Core component ablation. (b) Training paradigm comparison. (c) Distillation target variants. (d) Perturbation operator choices. (e) Attribution top-$k$ sensitivity. (f) Answer accuracy under component removal.}
  \label{fig:ablation_bar}
  \vspace{-0.5em}
\end{figure*}

\paragraph{Human study.}
On six benchmarks, we sample 150 image-cue pairs per cue source (GT, EDD, PATE-Forensics, ours), totaling 3{,}600 pairs. Each pair is evaluated by three of ten annotators experienced in forgery detection, and majority voting decides whether a cue is supported under the three criteria of App.~\ref{sec:cawe_eval}. As shown in Fig.~\ref{fig:human_observability}, GT cues reach support rates of 21\% to 51\% across benchmarks, while ours achieves 80\% to 90\%, outperforming EDD and PATE-Forensics.
We further verify that the cue-level supportability score $w_c = \frac{1}{|\operatorname{span}(c)|}\sum_{t\in\operatorname{span}(c)} w_{c,t}$ correlates with human judgment by sorting 1{,}000 FF++ cues into seven bins (Table~\ref{tab:supportability_validation}). Each cue receives a binary \emph{supported}/\emph{unsupported} label from the VLM evaluator and three human annotators via majority vote. VLM and human support rates rise with $w_c$, validating it as a supportability proxy.

\begin{table}[t]
\centering
\small
\resizebox{\columnwidth}{!}{%
\begin{tabular}{lccc}
\toprule
\textbf{Percentile $w_c$} & \textbf{\# Cues} & \textbf{VLM Sup. $\uparrow$} & \textbf{Human Sup. $\uparrow$} \\
\midrule
0\,--\,15\%   & 150 & \ffRank{A}{10.0} & \ffRank{A}{14.7} \\
15\,--\,30\%  & 150 & \ffRank{A}{24.0} & \ffRank{A}{28.7} \\
30\,--\,45\%  & 150 & \ffRank{A}{36.7} & \ffRank{A}{42.0} \\
45\,--\,60\%  & 150 & \ffRank{A}{50.7} & \ffRank{A}{54.7} \\
60\,--\,75\%  & 150 & \ffRank{B}{64.0} & \ffRank{B}{68.0} \\
75\,--\,90\%  & 150 & \ffRank{C}{76.7} & \ffRank{C}{78.7} \\
90\,--\,100\% & 100 & \ffRank{D}{88.0} & \ffRank{D}{86.0} \\
\bottomrule
\end{tabular}%
}
\caption{Supportability score validation on FF++ (1{,}000 cues). Higher $w_c$ bins yield monotonically higher VLM and human support rates (\%).}
\label{tab:supportability_validation}
\vspace{-1.5em}
\end{table}

\subsection{Qualitative results}
Fig.~\ref{fig:vis} compares evidence on two forged samples. M2F2-Det and Identity-Aware VLM misclassify the image as real, while FakeShield and VLFFD produce correct labels but resort to generic, template-like evidence that could apply to any forgery. Our method generates specific cues, such as the eyeglass-eye boundary inconsistency in the first sample and the skin-beard transition artifact in the second, both verifiable in the input image, while baseline cues stay vague and unlocalized. 
\subsection{Ablation Study}

Fig.~\ref{fig:ablation_bar} reports six ablation groups on FF++, Celeb-DF v2, and DFDC (FF++ summary in Fig.~\ref{fig:paradigm_comparison}). Removing TU, VS, or the evidence-first format (App.~\ref{sec:cue_record}) raises CAWE, and dropping both audits degrades the method to standard OPD with nearly tripled unfaithfulness, confirming both audits are indispensable (a). The same backbone and data trained with SFT, DPO, GRPO, or OPD yield scores above 58\%, so the gain stems from the audit rather than backbone or data (b). Among distillation targets, cue-masked distillation outperforms full-context, base-only, random-drop, and no-distillation variants, as it blocks literal cue leakage while keeping privileged guidance (c). Zero-out masking is the best perturbation, since noise and replace retain residual information and blur is too mild (d). The attribution scope $k{=}32$ strikes the best balance: smaller $k$ under-perturbs and larger $k$ corrupts irrelevant content (e). The full model also attains the highest ACC among all variants (f).

\section{Conclusion}

We trace the CAWE failure in forgery detection to pipeline-derived annotations, which introduce privileged cues unsupported by the image alone, and propose a dual supportability audit combining transfer utility and visual sensitivity that retains only visually supportable cues. Experiments across eight benchmarks show that our method substantially reduces CAWE while maintaining detection accuracy, and a human study confirms that the supportability score tracks human-judged support.

\section{Limitations}

The audit evaluates cues independently and can only filter teacher-provided candidates, so it may miss interacting or unannotated evidence. It also increases training cost and remains untested beyond facial forgeries and single-turn English explanations; broader modalities and settings require further study. The supportability scores depend on attribution quality and fixed hyperparameters, which may require recalibration across architectures or image resolutions. 

\bibliography{custom}

\clearpage
\onecolumn
\appendix
\pagestyle{plain}
\thispagestyle{plain}

\nolinenumbers
\vspace{0.6em}
\noindent\hspace*{-0.2em}{\normalfont\Large\bfseries Appendix\par}
\vspace{0.25em}
\noindent\hspace*{-0.2em}{\normalfont\large\bfseries Table of Contents\par}
\vspace{-0.08em}
\noindent\makebox[\linewidth]{\rule{\linewidth}{0.4pt}}\par

\makeatletter
\def\addcontentsline#1#2#3{%
  \addtocontents{#1}{\protect\contentsline{#2}{#3}{\thepage}%
    {\@currentHref}}}
\makeatother

\begingroup
\hypersetup{linkcolor=AppTocBlue}
\contentsmargin{3.0em}
\titlecontents{section}
  [2.1em]
  {\vspace{0.5em}\color{AppTocBlue}\normalsize\mdseries}
  {\contentslabel{2.4em}}
  {}
  {\titlerule*[0.45pc]{.}\contentspage}
\titlecontents{subsection}
  [5.4em]
  {\vspace{0.15em}\color{AppTocBlue}\small\mdseries}
  {\contentslabel{3.0em}}
  {}
  {\titlerule*[0.45pc]{.}\contentspage}
\titlecontents{subsubsection}
  [8.8em]
  {\vspace{0.08em}\color{AppTocBlue}\footnotesize\mdseries}
  {\contentslabel{3.6em}}
  {}
  {\titlerule*[0.45pc]{.}\contentspage}
\startcontents[sections]
\printcontents[sections]{l}{1}{\setcounter{tocdepth}{3}}
\vspace{0.08em}
\noindent\makebox[\linewidth]{\rule{\linewidth}{0.4pt}}\par
\endgroup
\clearpage
\twocolumn

\section{Proofs and Derivations}
\label{sec:proofs}

We justify the three components of the supportability criterion (Sec.~\ref{sec:formulation}).

\begin{proposition}[Transfer Utility measures direct-cue dependence]
\label{prop:tu}
Let $z_t$ denote the annotated cue token at position $t\in\operatorname{span}(c)$. The transfer utility is defined as
\begin{equation*}
\operatorname{TU}_{c,t}
=
\bigl[
\log p_{c}^{\mathrm{base},t}(z_t)
-
\log p_{c}^{\mathrm{mask},t}(z_t)
\bigr]_+,
\end{equation*}
where the full-context teacher $p_{c}^{\mathrm{base},t}$ observes the target cue and the cue-masked teacher $p_{c}^{\mathrm{mask},t}$ does not.

If the cue adds no predictive information, removing it leaves the prediction unchanged, i.e., the cue is inferable from the remaining context:
\begin{equation*}
p_{c}^{\mathrm{base},t}(z_t)
\approx
p_{c}^{\mathrm{mask},t}(z_t)
\;\;\Longrightarrow\;\;
\operatorname{TU}_{c,t}\approx 0.
\end{equation*}
If $z_t$ is likely only when the cue text is present, then
\begin{equation*}
p_{c}^{\mathrm{base},t}(z_t)
\gg
p_{c}^{\mathrm{mask},t}(z_t)
\;\;\Longrightarrow\;\;
\operatorname{TU}_{c,t}\gg 0,
\end{equation*}
indicating strong direct-cue dependence.

The text-side weight $g_{c,t}^{\mathrm{text}} = \exp(-\operatorname{TU}_{c,t}/\tau_a)$ satisfies
\begin{equation*}
\frac{\partial\, g_{c,t}^{\mathrm{text}}}
{\partial\, \operatorname{TU}_{c,t}}
=
-\frac{1}{\tau_a}\,g_{c,t}^{\mathrm{text}} < 0,
\end{equation*}
so tokens that copy rather than infer receive lower weight.
\end{proposition}

TU does not verify grounding in the current image: a low-TU cue may merely reflect a dataset-level pattern.

\begin{proposition}[Differential visual sensitivity removes generic perturbation effects]
\label{prop:vs}
The visual sensitivity test compares the cue likelihood degradation caused by perturbing high-attribution visual tokens versus low-attribution visual tokens of the same size $k$. Suppose the degradation for any token set $M$ decomposes as
\begin{equation*}
\Delta \ell_c(M)
=
\Delta_{\mathrm{generic}}
+
\Delta_{\mathrm{cue}}(M),
\end{equation*}
where $\Delta_{\mathrm{generic}}$ is the drop from removing any $k$ tokens and $\Delta_{\mathrm{cue}}(M)$ is the cue-specific contribution of $M$. Then
\begin{align*}
\Delta \ell_c(M_c^{\mathrm{high}})
&=
\Delta_{\mathrm{generic}}
+
\Delta_{\mathrm{cue}}(M_c^{\mathrm{high}}), \\
\Delta \ell_c(M_c^{\mathrm{low}})
&=
\Delta_{\mathrm{generic}}
+
\Delta_{\mathrm{cue}}(M_c^{\mathrm{low}}).
\end{align*}
Taking the difference cancels the shared generic term:
\begin{align*}
\operatorname{VS}(c)
&=
\bigl[\Delta \ell_c(M_c^{\mathrm{high}})
-
\Delta \ell_c(M_c^{\mathrm{low}})\bigr]_+ \\
&=
\bigl[\Delta_{\mathrm{cue}}(M_c^{\mathrm{high}})
-
\Delta_{\mathrm{cue}}(M_c^{\mathrm{low}})\bigr]_+.
\end{align*}
Thus VS tests whether the high-attribution region matters more to the cue than a same-size low-attribution region, rather than whether the model uses the image at all.

\emph{Remark.} The cancellation assumes matched generic effects between the two sets; this is approximate but reasonable given the identical zero-out operation and equal set size.
\end{proposition}

We next combine both audits into a single weight.

\begin{proposition}[The multiplicative joint gate is a soft relaxation of the supportability criterion]
\label{prop:gate}
Supportability requires low TU \emph{and} high VS:
\begin{equation*}
\operatorname{Supp}(c)
\;\Longleftrightarrow\;
\text{low TU}
\;\wedge\;
\text{high VS}.
\end{equation*}
Ignoring the span indicator, we implement it as
\begin{equation*}
w_{c,t}
=
g_{c,t}^{\mathrm{text}} \cdot g_c^{\mathrm{vis}},
\end{equation*}
where both factors lie in $[0,1]$. For any $a,b\in[0,1]$, we have $ab \leq \min(a,b)$, so
\begin{equation*}
w_{c,t}
\leq
\min\!\bigl(g_{c,t}^{\mathrm{text}},\; g_c^{\mathrm{vis}}\bigr).
\end{equation*}
Hence $w_{c,t}\to 0$ if either $g_{c,t}^{\mathrm{text}}\to 0$ (direct-cue dependence) or $g_c^{\mathrm{vis}}\to 0$ (no visual grounding), and $w_{c,t}\approx 1$ only when both are $\approx 1$. An additive form would let one dimension compensate for the other, violating the AND semantics.
\end{proposition}

\section{Evidence-first Single-cue Record}
\label{sec:cue_record}

Table~\ref{tab:info_access} summarizes the teacher-student information asymmetry. The teacher receives the full annotation $g=(y,o,r,c)$ and any available mask or source image; the student sees only $I$ and $q$. The cue-masked teacher (Sec.~\ref{sec:transfer}) additionally withholds $c$.

\begin{table}[t]
\centering
\small
\begin{tabular}{lcc}
\toprule
\textbf{Information} & \textbf{Teacher} & \textbf{Student} \\
\midrule
Image $I$            & \ding{51} & \ding{51} \\
Task prompt $q$      & \ding{51} & \ding{51} \\
GT label $y$         & \ding{51} & \ding{55} \\
Forgery type $o$     & \ding{51} & \ding{55} \\
Manipulation region $r$ & \ding{51} & \ding{55} \\
Target cue $c$       & Base: \ding{51} / Masked: \ding{55} & \ding{55} \\
Manipulation mask $m$ & if available & \ding{55} \\
Source image $I_{\mathrm{src}}$ & if available & \ding{55} \\
\bottomrule
\end{tabular}
\caption{Information accessibility for teacher and student.}
\label{tab:info_access}
\end{table}

Each training record $(I_i,q_i,g_i)$ carries the single annotated cue $c_i$ of $g_i$; an image annotated with several cues appears in several records, each carrying one cue. The record places the cue before the decision:
\begin{quote}
\texttt{Evidence:}\\
\texttt{- region: <region>}\\
\texttt{- cue: <cue>}\\
\texttt{Decision: <real/fake>}
\end{quote}
This makes the cue span stably locatable for token-level scoring, and the scoring prefix ends before \texttt{Decision:}, reducing post-hoc, answer-conditioned rationalization. The single-cue format applies only to training records; at inference, the student is free to list multiple \texttt{- region/- cue} entries under \texttt{Evidence:}, and produces about two cues per sample on average.

\section{CAWE Evaluation Protocol}
\label{sec:cawe_eval}

We measure explanation faithfulness with the Correct-Answer Wrong-Evidence (CAWE) metric, following the protocol of Kuckreja et al.~\cite{kuckreja2026pixels}.

\paragraph{Evidence evaluator.}
We fine-tune a VLM evaluator on image-cue pairs labeled \emph{supported} or \emph{unsupported}. A cue is unsupported if it fails \textbf{any} of three dimensions:
\begin{enumerate}[leftmargin=*,nosep]
  \item \textbf{Factual correctness}: the artifact is absent from the image.
  \item \textbf{Visual verifiability}: the artifact cannot be verified at the given resolution (e.g., ``subtle frequency-domain inconsistency'').
  \item \textbf{Spatial specificity}: the cue is too generic to localize (e.g., ``the image looks unnatural'').
\end{enumerate}
Negative pairs pair an image with a cue from another sample, a cue describing invisible artifacts, or an overly generic cue, violating dimensions 1, 2, and 3, respectively.

\paragraph{CAWE computation.}
Generated explanations are segmented into cues at the sentence level, each sentence stating one evidence claim. Among $N_{\mathrm{corr}}$ correctly classified samples, each generated cue is judged independently, and a sample counts as a CAWE case if \textbf{any} cue is unsupported. With $N_{\mathrm{unsup}}$ such samples,
\begin{equation*}
\mathrm{CAWE}
=
\frac{N_{\mathrm{unsup}}}{N_{\mathrm{corr}}}.
\end{equation*}
For correctly classified real images, stating that no artifact is observed is \emph{supported}, whereas hallucinating specific artifacts (e.g., ``slight skin smoothing around the cheeks'') is \emph{unsupported}. CAWE thus decouples faithfulness from detection accuracy.

\paragraph{Evaluator reliability.}
We compare the evaluator with human majority votes (three annotators) on the 3{,}600 image-cue pairs from the human study (Sec.~\ref{sec:experiments}), computing metrics on each 600-pair benchmark subset and macro-averaging over six benchmarks (Table~\ref{tab:evaluator_reliability}).

\begin{table}[t]
\centering
\small
\begin{tabular}{lccccc}
\toprule
 & \textbf{Acc} & \textbf{Prec.} & \textbf{Rec.} & \textbf{F1} & \textbf{$\kappa$} \\
\midrule
CAWE Eval. & 87.4 & 91.0 & 89.5 & 90.2 & 0.72 \\
\bottomrule
\end{tabular}
\caption{Reliability of the CAWE evaluator against human majority vote on 3{,}600 image-cue pairs. Metrics (\%) are macro-averaged over six benchmarks; $\kappa$ is Cohen's kappa.}
\label{tab:evaluator_reliability}
\end{table}

The evaluator reaches 87.4\% accuracy and $\kappa{=}0.72$ (substantial agreement) with balanced precision and recall, supporting it as a reliable proxy for human judgment.

\paragraph{Cue-multiplicity control.}
Since models producing more cues are more likely to incur a CAWE case, we retain one random cue per correctly classified sample and recompute CAWE, averaging over five seeds.

\begin{table*}[t]
\centering
\renewcommand{\arraystretch}{1.05}
\resizebox{\textwidth}{!}{%
\begin{tabular}{l cc cc cc cc cc}
\toprule
\textbf{Method}
& \multicolumn{2}{c}{\textbf{FF++}}
& \multicolumn{2}{c}{\textbf{Celeb-DF v2}}
& \multicolumn{2}{c}{\textbf{DFDC}}
& \multicolumn{2}{c}{\textbf{WildDF}}
& \multicolumn{2}{c}{\textbf{Average}} \\
\cmidrule(lr){2-3} \cmidrule(lr){4-5} \cmidrule(lr){6-7} \cmidrule(lr){8-9} \cmidrule(lr){10-11}
& Orig. & 1-Cue & Orig. & 1-Cue & Orig. & 1-Cue & Orig. & 1-Cue & Orig. & 1-Cue \\
\midrule
FFAA               & \ffRank{A}{62.8} & \ffRank{A}{49.8} & \celebRank{A}{68.4} & \celebRank{A}{54.2} & \dfdcRank{A}{65.2} & \dfdcRank{A}{51.7} & \wildRank{A}{71.6} & \wildRank{A}{57.4} & \averageRank{A}{67.0} & \averageRank{A}{53.3} \\
VLFFD              & \ffRank{A}{56.3} & \ffRank{A}{44.7} & \celebRank{A}{60.1} & \celebRank{A}{48.2} & \dfdcRank{A}{62.4} & \dfdcRank{A}{49.5} & \wildRank{A}{65.8} & \wildRank{A}{52.6} & \averageRank{A}{61.2} & \averageRank{A}{48.8} \\
Identity-Aware VLM & \ffRank{A}{53.6} & \ffRank{A}{41.8} & \celebRank{A}{55.4} & \celebRank{A}{43.1} & \dfdcRank{A}{61.7} & \dfdcRank{A}{48.7} & \wildRank{A}{59.2} & \wildRank{A}{46.0} & \averageRank{A}{57.5} & \averageRank{A}{44.9} \\
RLSBI              & \ffRank{A}{58.1} & \ffRank{A}{45.9} & \celebRank{A}{63.8} & \celebRank{A}{50.7} & \dfdcRank{A}{62.0} & \dfdcRank{A}{49.3} & \wildRank{A}{68.4} & \wildRank{A}{54.6} & \averageRank{A}{63.1} & \averageRank{A}{50.1} \\
SIDA               & \ffRank{A}{54.5} & \ffRank{A}{43.0} & \celebRank{A}{58.6} & \celebRank{A}{46.8} & \dfdcRank{A}{63.8} & \dfdcRank{A}{50.2} & \wildRank{A}{61.0} & \wildRank{A}{48.0} & \averageRank{A}{59.5} & \averageRank{A}{47.0} \\
FakeVLM            & \ffRank{A}{48.7} & \ffRank{A}{37.5} & \celebRank{A}{51.2} & \celebRank{A}{39.8} & \dfdcRank{A}{54.8} & \dfdcRank{A}{42.6} & \wildRank{A}{52.4} & \wildRank{A}{40.7} & \averageRank{A}{51.8} & \averageRank{A}{40.2} \\
M2F2-Det           & \ffRank{A}{50.4} & \ffRank{A}{38.4} & \celebRank{A}{52.1} & \celebRank{A}{40.1} & \dfdcRank{A}{58.6} & \dfdcRank{A}{45.9} & \wildRank{A}{55.3} & \wildRank{A}{42.7} & \averageRank{A}{54.1} & \averageRank{A}{41.8} \\
FakeShield         & \ffRank{A}{41.2} & \ffRank{A}{30.9} & \celebRank{A}{43.8} & \celebRank{A}{32.7} & \dfdcRank{A}{49.5} & \dfdcRank{A}{37.8} & \wildRank{A}{46.7} & \wildRank{A}{35.2} & \averageRank{A}{45.3} & \averageRank{A}{34.2} \\
EDD                & \ffRank{B}{38.5} & \ffRank{B}{29.4} & \celebRank{B}{40.2} & \celebRank{B}{30.8} & \dfdcRank{B}{42.6} & \dfdcRank{B}{32.7} & \wildRank{B}{39.8} & \wildRank{B}{30.2} & \averageRank{B}{40.3} & \averageRank{B}{30.8} \\
PATE-Forensics     & \ffRank{C}{36.4} & \ffRank{C}{26.9} & \celebRank{C}{35.6} & \celebRank{C}{26.4} & \dfdcRank{C}{38.2} & \dfdcRank{C}{28.6} & \wildRank{C}{36.5} & \wildRank{C}{27.0} & \averageRank{C}{36.7} & \averageRank{C}{27.2} \\
\midrule
\textbf{Ours}      & \ffRank{D}{21.6} & \ffRank{D}{11.5} & \celebRank{D}{22.8} & \celebRank{D}{12.1} & \dfdcRank{D}{27.4} & \dfdcRank{D}{14.8} & \wildRank{D}{24.1} & \wildRank{D}{12.9} & \averageRank{D}{24.0} & \averageRank{D}{12.8} \\
\bottomrule
\end{tabular}%
}
\caption{CAWE (\%) under the original evaluation and the one-cue control. ``1-Cue'' retains one uniformly sampled cue per correctly classified sample and averages results over five random seeds.}
\label{tab:one_cue_cawe}
\end{table*}

Controlling for cue multiplicity lowers CAWE for all methods, while ours remains best on every benchmark (Table~\ref{tab:one_cue_cawe}). The drop is larger for ours because it lists about two cues per sample, so its per-cue unsupported rate (12.8\% on average) is less than half that of the strongest baseline. This per-cue support also exceeds that of the top-$w_c$ training cues (Table~\ref{tab:supportability_validation}) because, for correctly classified real images, ``no visible artifact'' statements count as supported.

\section{Implementation Details}
\label{sec:impl_details}

\subsection{Training and Inference}

Algorithm~\ref{alg:training} summarizes training; Table~\ref{tab:impl_summary} lists settings not reported in the main text.

\begin{algorithm}[t]
\caption{Supportability-Aware Distillation}
\label{alg:training}
\small
\begin{algorithmic}[1]
\REQUIRE Image set $\mathcal{D}$; frozen teacher $p_T$; student $p_S$; TU temperature $\tau_a$; VS margin $m_{\mathrm{pert}}$, temperature $T_{\mathrm{pert}}$; top/bottom-$k$; loss weights $\lambda_{\mathrm{ans}}, \lambda_{\mathrm{sup}}$
\FOR{each mini-batch $\{(I_i, q_i, g_i)\}$ from $\mathcal{D}$}
  \STATE $\tilde{y}_{<t} \leftarrow$ student greedy rollout on $(I_i, q_i)$, with cue tokens teacher-forced
  \FOR{each sample $i$ in the mini-batch, with cue $c \leftarrow c_i$ from $g_i$}
    \STATE \textbf{// Transfer Utility (text-side audit)}
    \STATE $p_c^{\mathrm{base},t} \leftarrow p_T(\cdot \mid I_i, q_i, g_i, \tilde{y}_{<t})$
    \STATE $p_c^{\mathrm{mask},t} \leftarrow p_T(\cdot \mid I_i, q_i, g_i^{-c}, \tilde{y}_{<t})$
    \STATE $\mathrm{TU}_{c,t} \leftarrow [\log p_c^{\mathrm{base},t}(z_t) - \log p_c^{\mathrm{mask},t}(z_t)]_+$
    \STATE $g_{c,t}^{\mathrm{text}} \leftarrow \exp(-\mathrm{TU}_{c,t} / \tau_a)$
    \STATE \textbf{// Visual Sensitivity (image-side audit)}
    \STATE $A_{c,u} \leftarrow \|\frac{\partial \ell_c}{\partial v_u} \odot v_u\|_1$ \quad (grad $\times$ activation)
    \STATE $M_c^{\mathrm{high}}, M_c^{\mathrm{low}} \leftarrow$ top-$k$, bottom-$k$ by $A_{c,u}$
    \STATE $\operatorname{VS}(c) \leftarrow [\ell_c(\mathcal{T}(I_i, M_c^{\mathrm{low}})) - \ell_c(\mathcal{T}(I_i, M_c^{\mathrm{high}}))]_+$
    \STATE $g_c^{\mathrm{vis}} \leftarrow \sigma((\operatorname{VS}(c) - m_{\mathrm{pert}}) / T_{\mathrm{pert}})$
    \STATE \textbf{// Joint weight}
    \STATE $w_{c,t} \leftarrow \mathbf{1}[t \in \operatorname{span}(c)] \cdot g_{c,t}^{\mathrm{text}} \cdot g_c^{\mathrm{vis}}$
  \ENDFOR
  \STATE $\mathcal{L}_{\mathrm{sup}} \leftarrow \frac{\sum_c \sum_t w_{c,t}\, D_{\mathrm{KL}}(\mathrm{sg}[p_c^{\mathrm{mask},t}] \| p_S^t)}{\sum_c \sum_t w_{c,t} + \epsilon}$
  \STATE $\mathcal{L}_{\mathrm{total}} \leftarrow \lambda_{\mathrm{ans}}\, \mathrm{CE}(y_i, p_S^{\mathrm{ans}}) + \lambda_{\mathrm{sup}}\, \mathcal{L}_{\mathrm{sup}}$
  \STATE Update $p_S$ via $\nabla \mathcal{L}_{\mathrm{total}}$
\ENDFOR
\end{algorithmic}
\end{algorithm}

Teacher and student are both initialized from \texttt{Qwen/Qwen3-VL-8B} without prior SFT; the teacher stays frozen. At each step, the student's greedy rollout prefix is fed to the full-context and cue-masked teachers in one batched forward pass. TU is computed online since $\tilde{y}_{<t}$ evolves; VS is deterministic per (image, cue) pair but computed jointly with TU for simplicity. We select the checkpoint with the lowest CAWE on a held-out 10\% split.

For VS (Algorithm~\ref{alg:training}, lines 10--13), gradient$\times$activation attribution is computed on the last vision-encoder hidden state w.r.t.\ the cue-masked log-likelihood and reduced per token via the $L_1$ norm; top-$k$ and bottom-$k$ sets have equal size, and each cue is audited independently. At inference, the decision is parsed from the ``Decision:'' tag (unparseable outputs count as incorrect), and AUC/EER use $s_{\mathrm{fake}} = p(\text{Fake}) / (p(\text{Fake}) + p(\text{Real}))$ at the decision position.

\begin{table}[t]
\centering
\small
\begin{tabular}{@{}lp{4.2cm}@{}}
\toprule
\textbf{Setting} & \textbf{Value} \\
\midrule
Optimizer & AdamW \\
Weight decay & 0.01 \\
Scheduler & Cosine, 3\% warmup \\
Grad.\ clipping & Max norm 1.0 \\
Precision & BF16 \\
Max seq.\ length & 2048 tokens \\
Image resolution & 448 px (longer side) \\
Image processor & Qwen3-VL default \\
Fine-tuning & Full (no LoRA) \\
Random seed & 42 \\
Ckpt.\ selection & Lowest val.\ CAWE (10\%) \\
\midrule
Teacher / Student & \texttt{Qwen/Qwen3-VL-8B} \\
Teacher update & Frozen \\
TU/VS computation & Online (every step) \\
Attribution layer & Last vision hidden state \\
Attribution reduction & $L_1$(grad $\times$ activation) \\
Perturbation & Zero-out embedding \\
\midrule
Decoding & Greedy \\
Max new tokens & 1024 \\
AUC/EER score & $\frac{p(\text{F})}{p(\text{F})+p(\text{R})}$ \\
Decision parsing & ``Decision:'' tag \\
\bottomrule
\end{tabular}
\caption{Implementation summary. Main-text hyperparameters ($\tau_a$, $m_{\mathrm{pert}}$, $T_{\mathrm{pert}}$, $k$, $\lambda_{\mathrm{ans}}$, $\lambda_{\mathrm{sup}}$, epochs, LR, batch size, hardware) are omitted.}
\label{tab:impl_summary}
\end{table}

\subsection{Cue Extraction Pipeline}
\label{sec:cue_pipeline}

The cue $c$ in $g=(y,o,r,c)$ is derived from native dataset metadata in two stages.

\paragraph{Stage 1: Rule-based extraction.}
Per-dataset templates convert metadata (forgery method labels, region masks, identity pairs) into standardized cue phrases; e.g., ``NeuralTextures'' with a mouth mask yields ``neural texture re-rendering in the mouth region''. This covers most samples.

\paragraph{Stage 2: VLM-based refinement.}
For less structured annotations (e.g., ForgeryNet free-form descriptions) or overly generic template outputs, GPT-4o distills the metadata into one concise cue, normalized to the rule-based vocabulary. Over 95\% of 200 randomly sampled outputs correctly reflect the described forgery.

Neither stage checks whether a cue is supportable by the image; this is the role of the supportability audit (Secs.~\ref{sec:transfer}--\ref{sec:distill}).


\section{Supportability Analysis}
\label{sec:supportability_analysis}

We show that visual supportability varies across datasets (Sec.~\ref{sec:dataset_visual_stats}) and that the gains stem from the audit rather than teacher capacity (Sec.~\ref{sec:teacher_robustness}).

\subsection{Dataset Visual Statistics}
\label{sec:dataset_visual_stats}

We manually inspect a random sample of test images per dataset to check whether annotated cues have visible evidence (Table~\ref{tab:dataset_visual_stats}); DeepFakeBench is included for analysis only.

\begin{table}[t]
\centering
\resizebox{\columnwidth}{!}{%
\small
\begin{tabular}{lccc}
\toprule
\textbf{Dataset} & \textbf{Resolution} & \textbf{Vis. Sup. (\%)} & \textbf{\#Sampled} \\
\midrule
FF++~\cite{rossler2019faceforensics++}           & $256{\times}256$  & 51.6 & 200 \\
Celeb-DF v2~\cite{li2020celeb}    & $256{\times}256$  & 40.3 & 160 \\
DFDC~\cite{dolhansky2020deepfake}           & $320{\times}320$  & 30.5 & 240 \\
WildDF~\cite{zi2020wilddeepfake}         & varies            & 21.4 & 180 \\
DeeperForensics~\cite{jiang2020deeperforensics} & $256{\times}256$ & 35.2 & 200 \\
ForgeryNet~\cite{he2021forgerynet}     & $256{\times}256$  & 25.7 & 220 \\
Celeb-DF++~\cite{li2507celeb}     & $256{\times}256$  & 42.5 & 170 \\
DF40~\cite{yan2024df40}           & $256{\times}256$  & 38.2 & 200 \\
DeepFakeBench~\cite{yan2023deepfakebench}  & $256{\times}256$  & 34.8 & 190 \\
\bottomrule
\end{tabular}%
}
\caption{Per-dataset image resolution and visual-support ratio. ``Vis. Sup.'' denotes the percentage, over all annotated cues on the sampled images, of cues whose visual evidence is identifiable by human inspection. \#Sampled is the number of randomly sampled test images inspected.}
\label{tab:dataset_visual_stats}
\end{table}

FF++ and the Celeb-DF series show the highest supportability (40.3--51.6\%), DeeperForensics and multi-method benchmarks (DF40, DeepFakeBench) are moderate (34.8--38.2\%), DFDC, with diverse capture conditions and perturbations, is lower (30.5\%), and ForgeryNet and WildDF, with diverse compression and resolution, are lowest (25.7\% and 21.4\%). Without the VS gate, ungrounded cues would be distilled equally, inflating CAWE on harder benchmarks.

\subsection{Teacher Robustness}
\label{sec:teacher_robustness}

To test whether the high CAWE of Standard OPD merely reflects limited teacher capacity, we fix the student (Qwen3-VL-8B) and vary the teacher across six models from three families (3B--38B) on FF++.

\begin{table}[t]
\centering
\resizebox{\columnwidth}{!}{%
\small
\begin{tabular}{llccc}
\toprule
\textbf{Method} & \textbf{Teacher} & \textbf{Params} & \textbf{ACC} & \textbf{CAWE} \\
\midrule
\multicolumn{5}{l}{\emph{Qwen2.5-VL family}~\cite{bai2025qwen25vltechnicalreport}} \\
Standard OPD & Qwen2.5-VL-3B  & 3B  & \ffRank{A}{95.40} & \ffRank{A}{61.4} \\
Standard OPD & Qwen2.5-VL-7B  & 7B  & \ffRank{A}{96.20} & \ffRank{A}{59.3} \\
\midrule
\multicolumn{5}{l}{\emph{Qwen3-VL family}~\cite{bai2025qwen3}} \\
Standard OPD & Qwen3-VL-8B    & 8B  & \ffRank{A}{96.50} & \ffRank{A}{58.6} \\
Standard OPD & Qwen3-VL-32B   & 32B & \ffRank{B}{97.30} & \ffRank{B}{54.9} \\
\midrule
\multicolumn{5}{l}{\emph{InternVL3 family}~\cite{zhu2025internvl3}} \\
Standard OPD & InternVL3-8B  & 8B  & \ffRank{A}{96.35} & \ffRank{A}{58.0} \\
Standard OPD & InternVL3-38B & 38B & \ffRank{C}{97.50} & \ffRank{C}{54.1} \\
\midrule
Ours         & Qwen3-VL-8B    & 8B  & \ffRank{D}{99.12} & \ffRank{D}{21.6} \\
\bottomrule
\end{tabular}%
}
\caption{Effect of teacher identity and scale on FF++ (student: Qwen3-VL-8B; same protocol as Table~\ref{tab:main}).}
\label{tab:teacher_scale}
\end{table}

\begin{figure*}[t]
  \centering
  \includegraphics[width=\textwidth]{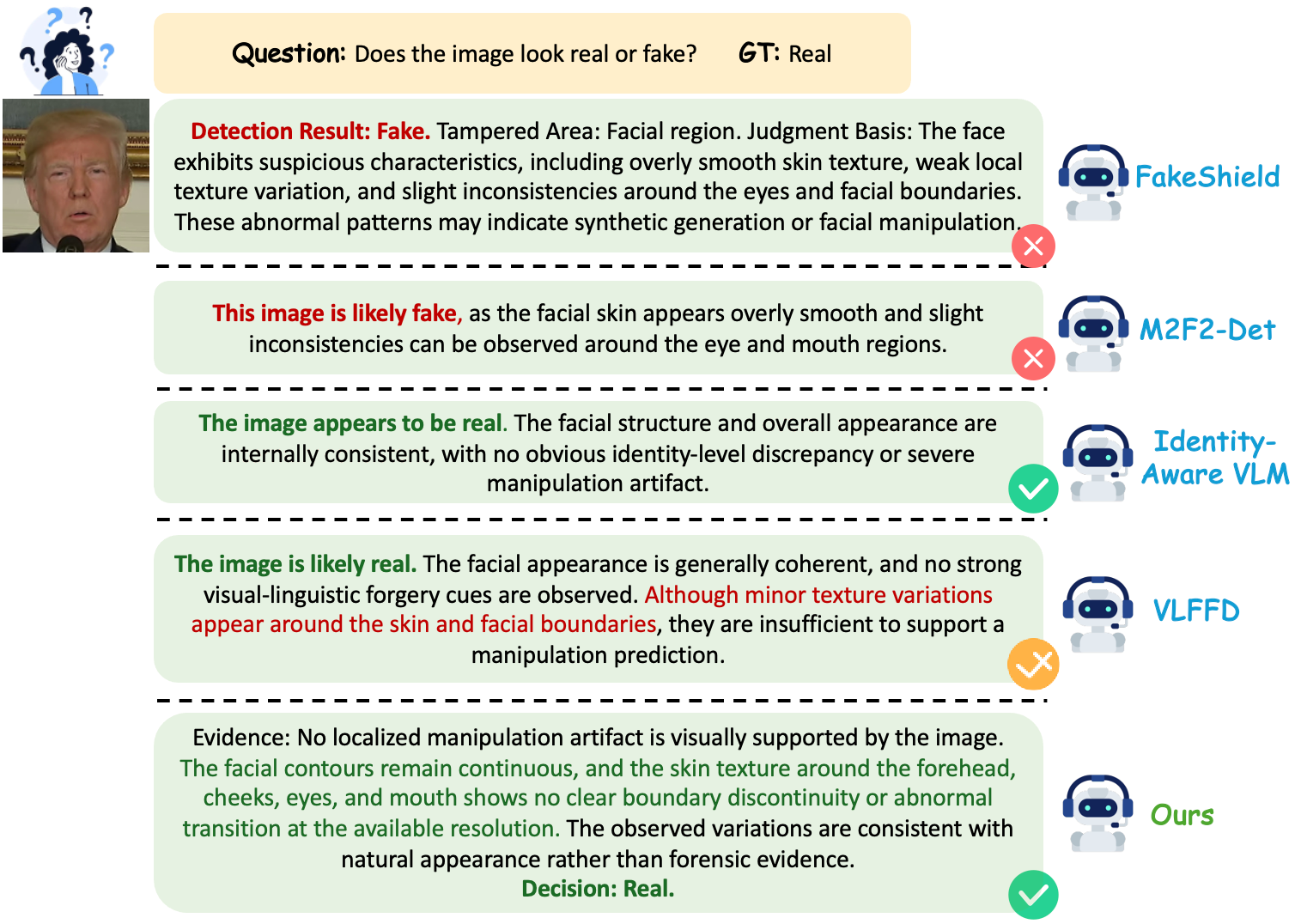}
  \caption{Visualization on a real image (Case 1). \textcolor{red}{$\times$}: wrong prediction; \textcolor{orange}{$\checkmark\!\times$}: correct prediction with unfaithful evidence; \textcolor{green!60!black}{$\checkmark$}: correct prediction with faithful evidence.}
  \label{fig:qualitative}
\end{figure*}

As shown in Table~\ref{tab:teacher_scale}, scaling the teacher within a family reduces CAWE by only 2--4 points; even InternVL3-38B retains 54.1\% CAWE, and 7B/8B-class teachers across families stay within 58--60\%. Ours with the default 8B teacher reaches 99.12\% ACC and 21.6\% CAWE, over 32 points lower than any Standard OPD setting. The bottleneck is thus unsupported privileged cues, not teacher capacity.

\section{Qualitative Examples}
\label{sec:qualitative}

\subsection{Comparison with Baselines}
\label{sec:qualitative_baselines}

We compare with recent explainable detectors on a real and a forged image.

\paragraph{Case 1: Real image (Fig.~\ref{fig:qualitative}).}
FakeShield and M2F2-Det misclassify it as \emph{Fake}, hallucinating ``overly smooth'' skin and inconsistencies around the eyes, mouth, or facial boundaries (Dimension 1). Identity-Aware VLM predicts \emph{Real} without fabricating specific artifacts, which counts as faithful (Appendix~\ref{sec:cawe_eval}); VLFFD also predicts \emph{Real} but still mentions minor texture variations near the facial boundaries (Dimension 1). Ours predicts \emph{Real} with a spatially grounded negative explanation (continuous cheek and jawline contours, no texture discontinuity) that a human can verify in the cited regions.

\begin{figure*}[t]
  \centering
  \includegraphics[width=\textwidth]{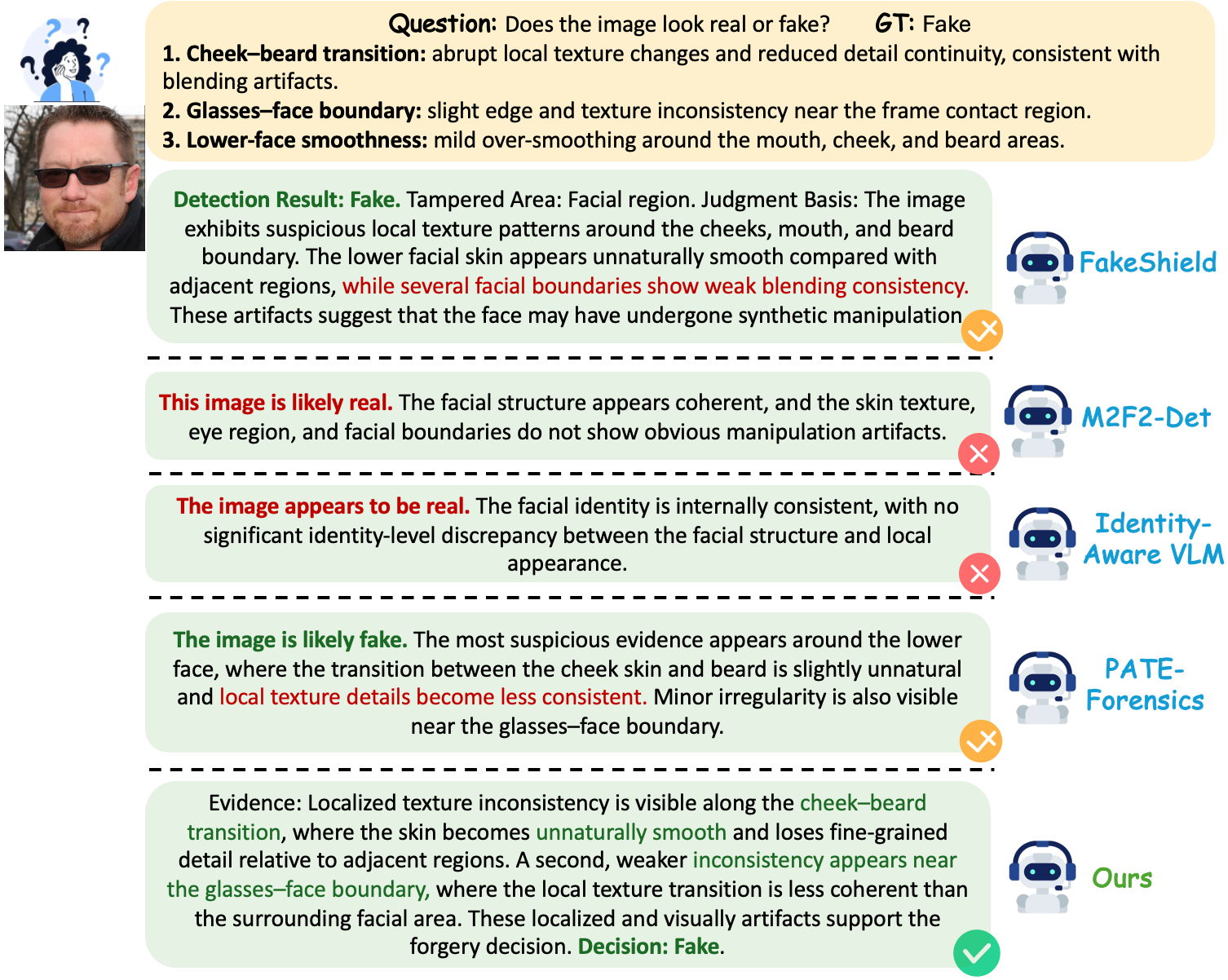}
  \caption{Visualization on a forged image (Case 2). Symbols follow Fig.~\ref{fig:qualitative}.}
  \label{fig:qualitative_fake}
\end{figure*}

\paragraph{Case 2: Forged image (Fig.~\ref{fig:qualitative_fake}).}
Artifacts identified by annotators: cheek-beard texture transition, glasses-face boundary irregularity, and lower-face over-smoothing. FakeShield detects the forgery but also cites an invisible ``weak blending consistency'' (Dimension 1). M2F2-Det and Identity-Aware VLM predict \emph{Real}. PATE-Forensics localizes the cheek-beard and glasses-face artifacts but adds ``local texture details become less consistent'' without saying where (Dimension 3). Ours pinpoints the cheek-beard texture inconsistency and the glasses-face boundary, each with an explicit region and observable pattern, satisfying all three dimensions.

\subsection{Before vs.\ After Training}
\label{sec:qualitative_training}

\begin{figure*}[t]
  \centering
  \includegraphics[width=\textwidth]{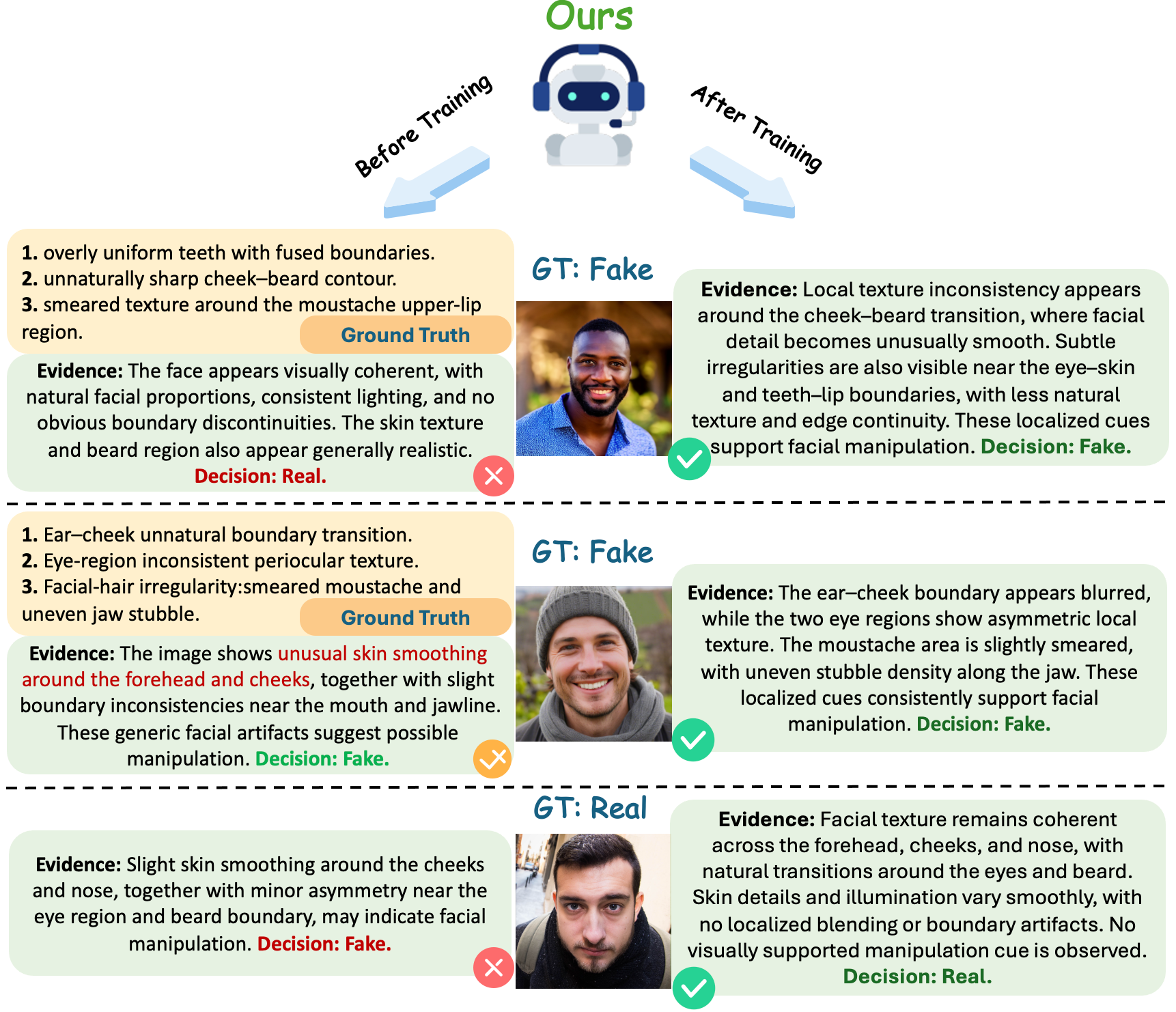}
  \caption{Before vs.\ after supportability-aware training. Left: explanations before training; right: explanations after training. \textcolor{red}{$\times$}: wrong prediction; \textcolor{orange}{$\checkmark\!\times$}: correct prediction with unfaithful evidence; \textcolor{green!60!black}{$\checkmark$}: correct prediction with faithful evidence.}
  \label{fig:qualitative_training}
\end{figure*}

Fig.~\ref{fig:qualitative_training} compares our model's explanations before and after training.

\paragraph{Sample 1 (GT: Fake).}
Before training, the model wrongly predicts \emph{Real} (``visually coherent \ldots\ no obvious boundary discontinuities''). After training, it predicts \emph{Fake} and localizes inconsistencies at the cheek-beard, eye-skin, and teeth-lip boundaries, two of which (cheek-beard, teeth-lip) match the artifacts identified by annotators (fused teeth, sharp cheek-beard contour, smeared moustache).

\paragraph{Sample 2 (GT: Fake).}
Both models predict \emph{Fake}, but before training the evidence is generic (``skin smoothing around the forehead and cheeks''). After training, the model pinpoints the blurred ear-cheek boundary, asymmetric periocular texture, and uneven jaw stubble, all identified by annotators.

\paragraph{Sample 3 (GT: Real).}
Before training, the model wrongly predicts \emph{Fake}, citing cheek smoothing and eye-beard asymmetry. After training, it predicts \emph{Real}, noting coherent texture and natural transitions with no supported manipulation cue.

Overall, training shifts the model from generic or hallucinated cues to localized, verifiable evidence.

\section{Extended Comparison}
\label{sec:extended_comparison}

Table~\ref{tab:extended} extends the comparison to DeeperForensics~\cite{jiang2020deeperforensics}, Celeb-DF v2~\cite{li2020celeb}, ForgeryNet~\cite{he2021forgerynet}, Celeb-DF++~\cite{li2507celeb}, and DF40~\cite{yan2024df40}, adding nine traditional and five explainable baselines, under the same multi-dataset protocol as Table~\ref{tab:main}. Tables~\ref{tab:paradigm} and~\ref{tab:extended} use the same checkpoint, so our results on DeeperForensics, ForgeryNet, and Celeb-DF++ coincide. For Celeb-DF v2, since Celeb-DF++ extends it and shares source videos, Table~\ref{tab:paradigm} evaluates on the test subset disjoint from Celeb-DF++ (99.10), whereas Tables~\ref{tab:main} and~\ref{tab:extended} use the full official test set (99.35).

\begin{table*}[t]
\centering
\renewcommand{\arraystretch}{1.05}
\resizebox{\textwidth}{!}{%
\begin{tabular}{ll cc cc cc cc cc cc cc}
\toprule
& & \multicolumn{2}{c}{\textbf{DeeperForensics}} & \multicolumn{2}{c}{\textbf{Celeb-DF v2}} & \multicolumn{2}{c}{\textbf{ForgeryNet}} & \multicolumn{2}{c}{\textbf{Celeb-DF++}} & \multicolumn{2}{c}{\textbf{DF40}} \\
\cmidrule(lr){3-4} \cmidrule(lr){5-6} \cmidrule(lr){7-8} \cmidrule(lr){9-10} \cmidrule(lr){11-12}
\textbf{Category} & \textbf{Method} & ACC & AUC & ACC & AUC & ACC & AUC & ACC & AUC & ACC & AUC \\
\midrule
\multirow{9}{*}{Traditional}
& EfficientNet~\cite{tan2019efficientnet} & \deeperRank{A}{88.60} & \deeperRank{A}{92.40} & \celebRank{A}{90.80} & \celebRank{A}{93.20} & \forgeryRank{A}{70.40} & \forgeryRank{A}{76.80} & \celebPlusRank{A}{89.60} & \celebPlusRank{A}{91.80} & \dfFortyRank{A}{72.50} & \dfFortyRank{A}{78.40} \\
& Face X-ray~\cite{li2020face} & \deeperRank{A}{90.40} & \deeperRank{A}{93.80} & \celebRank{A}{92.60} & \celebRank{A}{94.50} & \forgeryRank{A}{72.80} & \forgeryRank{A}{78.60} & \celebPlusRank{A}{91.80} & \celebPlusRank{A}{93.20} & \dfFortyRank{A}{74.20} & \dfFortyRank{A}{80.60} \\
& F3Net~\cite{qian2020thinking} & \deeperRank{A}{91.20} & \deeperRank{A}{94.60} & \celebRank{A}{93.40} & \celebRank{A}{95.80} & \forgeryRank{A}{74.50} & \forgeryRank{A}{80.40} & \celebPlusRank{A}{92.60} & \celebPlusRank{A}{94.50} & \dfFortyRank{A}{75.80} & \dfFortyRank{A}{82.40} \\
& MAT~\cite{zhao2021multi} & \deeperRank{A}{91.80} & \deeperRank{A}{95.10} & \celebRank{A}{93.80} & \celebRank{A}{96.20} & \forgeryRank{A}{75.20} & \forgeryRank{A}{81.60} & \celebPlusRank{A}{93.00} & \celebPlusRank{A}{94.80} & \dfFortyRank{A}{76.40} & \dfFortyRank{A}{83.20} \\
& SPSL~\cite{liu2021spatial} & \deeperRank{A}{90.80} & \deeperRank{A}{94.20} & \celebRank{A}{93.20} & \celebRank{A}{95.60} & \forgeryRank{A}{73.80} & \forgeryRank{A}{79.80} & \celebPlusRank{A}{92.40} & \celebPlusRank{A}{94.20} & \dfFortyRank{A}{75.20} & \dfFortyRank{A}{81.80} \\
& RECCE~\cite{cao2022end} & \deeperRank{A}{92.40} & \deeperRank{A}{95.60} & \celebRank{A}{94.20} & \celebRank{A}{96.50} & \forgeryRank{A}{76.80} & \forgeryRank{A}{83.20} & \celebPlusRank{A}{93.60} & \celebPlusRank{A}{95.40} & \dfFortyRank{A}{77.60} & \dfFortyRank{A}{84.50} \\
& IID~\cite{dong2023implicit} & \deeperRank{A}{92.80} & \deeperRank{A}{95.90} & \celebRank{A}{94.80} & \celebRank{A}{96.70} & \forgeryRank{A}{76.20} & \forgeryRank{A}{82.50} & \celebPlusRank{A}{93.80} & \celebPlusRank{A}{95.60} & \dfFortyRank{A}{78.40} & \dfFortyRank{A}{85.20} \\
& UCF~\cite{yan2023ucf} & \deeperRank{A}{93.40} & \deeperRank{A}{96.20} & \celebRank{A}{95.20} & \celebRank{A}{97.10} & \forgeryRank{A}{77.80} & \forgeryRank{A}{84.40} & \celebPlusRank{A}{94.40} & \celebPlusRank{A}{96.00} & \dfFortyRank{A}{79.60} & \dfFortyRank{A}{86.40} \\
& TALL~\cite{xu2023tall} & \deeperRank{A}{93.80} & \deeperRank{A}{96.50} & \celebRank{A}{95.60} & \celebRank{A}{97.40} & \forgeryRank{A}{78.40} & \forgeryRank{A}{85.20} & \celebPlusRank{A}{94.80} & \celebPlusRank{A}{96.40} & \dfFortyRank{A}{80.20} & \dfFortyRank{A}{87.10} \\
\midrule
\multirow{5}{*}{Explainable}
& RAIDx~\cite{li2025raidx} & \deeperRank{A}{91.60} & \deeperRank{A}{95.20} & \celebRank{A}{93.80} & \celebRank{A}{96.10} & \forgeryRank{A}{75.40} & \forgeryRank{A}{81.80} & \celebPlusRank{A}{93.20} & \celebPlusRank{A}{95.40} & \dfFortyRank{A}{76.80} & \dfFortyRank{A}{83.60} \\
& So-Fake~\cite{huang2025so} & \deeperRank{A}{92.40} & \deeperRank{A}{95.80} & \celebRank{A}{94.20} & \celebRank{A}{96.40} & \forgeryRank{A}{76.80} & \forgeryRank{A}{83.20} & \celebPlusRank{A}{93.60} & \celebPlusRank{A}{95.80} & \dfFortyRank{A}{77.40} & \dfFortyRank{A}{84.20} \\
& REVEAL~\cite{praharaj2025reveal} & \deeperRank{A}{93.80} & \deeperRank{A}{96.40} & \celebRank{A}{95.60} & \celebRank{A}{97.20} & \forgeryRank{A}{78.60} & \forgeryRank{A}{84.80} & \celebPlusRank{A}{94.80} & \celebPlusRank{A}{96.60} & \dfFortyRank{A}{79.20} & \dfFortyRank{A}{85.80} \\
& FP-DAPO~\cite{zhuforensic} & \deeperRank{B}{94.60} & \deeperRank{B}{97.10} & \celebRank{B}{96.20} & \celebRank{B}{97.80} & \forgeryRank{B}{80.40} & \forgeryRank{B}{86.60} & \celebPlusRank{B}{95.60} & \celebPlusRank{B}{97.20} & \dfFortyRank{B}{81.20} & \dfFortyRank{B}{88.00} \\
& GE-AIGC~\cite{guo2026grounding} & \deeperRank{C}{95.20} & \deeperRank{C}{97.40} & \celebRank{C}{96.80} & \celebRank{C}{98.10} & \forgeryRank{C}{81.60} & \forgeryRank{C}{87.40} & \celebPlusRank{C}{96.20} & \celebPlusRank{C}{97.60} & \dfFortyRank{C}{82.40} & \dfFortyRank{C}{88.80} \\
\midrule
& \textbf{Ours} & \deeperRank{D}{97.25} & \deeperRank{D}{98.80} & \celebRank{D}{99.35} & \celebRank{D}{99.86} & \forgeryRank{D}{90.80} & \forgeryRank{D}{95.60} & \celebPlusRank{D}{98.85} & \celebPlusRank{D}{99.40} & \dfFortyRank{D}{91.40} & \dfFortyRank{D}{96.20} \\
\bottomrule
\end{tabular}%
}
\caption{Extended comparison on five benchmarks with methods not included in Table~\ref{tab:main}. All models are trained on the combined multi-dataset split, following the same protocol as Table~\ref{tab:main}.}
\label{tab:extended}
\end{table*}

\begin{table}[t]
\centering
\renewcommand{\arraystretch}{1.05}
\resizebox{\columnwidth}{!}{%
\begin{tabular}{ll cc}
\toprule
\textbf{Category} & \textbf{Method} & CAWE$\downarrow$ (Avg.) & $\Delta$ vs.\ Ours \\
\midrule
\multirow{5}{*}{Explainable}
& RAIDx~\cite{li2025raidx} & \averageRank{A}{58.6} & +27.4 \\
& So-Fake~\cite{huang2025so} & \averageRank{A}{56.4} & +25.2 \\
& REVEAL~\cite{praharaj2025reveal} & \averageRank{A}{52.8} & +21.6 \\
& FP-DAPO~\cite{zhuforensic} & \averageRank{B}{48.6} & +17.4 \\
& GE-AIGC~\cite{guo2026grounding} & \averageRank{C}{46.8} & +15.6 \\
\midrule
& \textbf{Ours} & \averageRank{D}{31.2} & -- \\
\bottomrule
\end{tabular}%
}
\caption{Average CAWE across the five benchmarks for explainable methods. $\Delta$ denotes the gap relative to our method.}
\label{tab:extended_cawe}
\end{table}

Recent traditional detectors (UCF, TALL, RECCE) outperform earlier ones, yet all drop sharply on DF40 (40 forgery methods). The additional explainable methods keep high average CAWE (46.8--58.6\%, Table~\ref{tab:extended_cawe}), consistent with Table~\ref{tab:main}. Ours achieves the best ACC and AUC on all five benchmarks among the methods in Table~\ref{tab:extended} (on Celeb-DF v2, its AUC of 99.86 still trails M2F2-Det's 99.92 in Table~\ref{tab:main}); on DF40 it exceeds GE-AIGC by 9.0 ACC and 7.4 AUC points, with 15.6 points lower average CAWE. Our per-benchmark CAWE is 28.6 (DeeperForensics), 22.8 (Celeb-DF v2, full test set), 38.4 (ForgeryNet), 30.8 (Celeb-DF++), and 35.4 (DF40), averaging 31.2.

\section{Inference Efficiency}
\label{sec:inference_efficiency}

Table~\ref{tab:inference_efficiency} compares the inference cost of all methods in Table~\ref{tab:main} on a single A100 80GB GPU (batch size 1). Unlike Inf.\ in Table~\ref{tab:paradigm}, which counts all prompt, visual, and output tokens, Out Tok.\ counts only generated tokens.

\begin{table}[t]
\centering
\renewcommand{\arraystretch}{1.05}
\resizebox{\columnwidth}{!}{%
\begin{tabular}{llrccc}
\toprule
\textbf{Cat.} & \textbf{Method} & \textbf{Params} & \textbf{Out Tok.} & \textbf{Latency (s)} & \textbf{Thru. (samples/s)} \\
\midrule
\multirow{3}{*}{Traditional}
& Xception & 23M & -- & 0.008 & 125.0 \\
& SBI & 23M & -- & 0.012 & 83.3 \\
& Forensics Adapter & 151M & -- & 0.015 & 66.7 \\
\midrule
\multirow{10}{*}{Explainable}
& FFAA & 8B & \ffRank{A}{286} & \ffRank{A}{3.82} & \ffRank{A}{0.26} \\
& VLFFD & 8B & \ffRank{A}{245} & \ffRank{A}{3.28} & \ffRank{A}{0.30} \\
& Identity-Aware VLM & 8B & \ffRank{A}{312} & \ffRank{A}{4.16} & \ffRank{A}{0.24} \\
& RLSBI & 8B & \ffRank{A}{268} & \ffRank{A}{3.58} & \ffRank{A}{0.28} \\
& SIDA & 8B & \ffRank{A}{254} & \ffRank{A}{3.40} & \ffRank{A}{0.29} \\
& FakeVLM & 8B & \ffRank{B}{238} & \ffRank{B}{3.18} & \ffRank{B}{0.31} \\
& M2F2-Det & 8B & \ffRank{A}{302} & \ffRank{A}{4.04} & \ffRank{A}{0.25} \\
& FakeShield & 13B & \ffRank{A}{275} & \ffRank{A}{5.86} & \ffRank{A}{0.17} \\
& EDD & 8B & \ffRank{C}{224} & \ffRank{C}{2.98} & \ffRank{C}{0.34} \\
& PATE-Forensics & 8B & \ffRank{A}{318} & \ffRank{A}{4.24} & \ffRank{A}{0.24} \\
\midrule
& \textbf{Ours} & 8B & \ffRank{D}{156} & \ffRank{D}{2.08} & \ffRank{D}{0.48} \\
\bottomrule
\end{tabular}%
}
\caption{Inference efficiency comparison on FF++. Params: model parameters. Out Tok.: average output tokens. Latency: seconds per sample. Thru.: throughput (samples/second). Traditional methods do not generate text.}
\label{tab:inference_efficiency}
\end{table}

Traditional detectors are orders of magnitude faster but give no explanations. Among VLM-based methods, latency is dominated by output length; FakeShield is slowest (5.86s) due to its 13B backbone. Ours generates the fewest tokens (156, 30--51\% fewer than other 8B methods), yielding the lowest latency (2.08s, 30.2\% below EDD) and highest throughput, as down-weighting unsupported cues teaches concise explanations. Inf.\ in Table~\ref{tab:paradigm} (e.g., 547 on DeeperForensics) additionally includes $\approx$390 prompt and visual tokens. Table~\ref{tab:efficiency_breakdown} splits latency into visual encoding and text generation.

\begin{table}[t]
\centering
\renewcommand{\arraystretch}{1.05}
\resizebox{\columnwidth}{!}{%
\begin{tabular}{lcccc}
\toprule
\textbf{Method} & \textbf{Vis.\ Enc.\ (s)} & \textbf{Text Gen.\ (s)} & \textbf{Total (s)} & \textbf{Gen.\ Ratio} \\
\midrule
FFAA & 0.42 & \ffRank{A}{3.40} & \ffRank{A}{3.82} & \ffRank{A}{89.0\%} \\
VLFFD & 0.42 & \ffRank{B}{2.86} & \ffRank{B}{3.28} & \ffRank{B}{87.2\%} \\
EDD & 0.42 & \ffRank{C}{2.56} & \ffRank{C}{2.98} & \ffRank{C}{85.9\%} \\
PATE-Forensics & 0.42 & \ffRank{A}{3.82} & \ffRank{A}{4.24} & \ffRank{A}{90.1\%} \\
FakeShield & 0.48 & \ffRank{A}{5.38} & \ffRank{A}{5.86} & \ffRank{A}{91.8\%} \\
\textbf{Ours} & 0.42 & \ffRank{D}{1.66} & \ffRank{D}{2.08} & \ffRank{D}{79.8\%} \\
\bottomrule
\end{tabular}%
}
\caption{Inference time breakdown. Vis.\ Enc.: visual encoding time. Text Gen.: autoregressive text generation time. Gen.\ Ratio: fraction of total time spent on text generation.}
\label{tab:efficiency_breakdown}
\end{table}

Text generation accounts for 80--92\% of inference time. Ours reduces it to 1.66s (vs.\ 2.56s for EDD) purely through training, without inference-time pruning or early stopping.

\section{Extended Limitations and Future Work}
\label{sec:limitations}

\paragraph{Audit granularity and teacher dependence.}
Cues are scored independently, which may underestimate jointly supportable combinations (e.g., a blending boundary visible only with a color shift); group-level audits or cue graphs could help. The framework also only filters cues, so artifacts omitted by the teacher annotation cannot be learned; retrieval-augmented or open-vocabulary cue proposal could complement it.

\paragraph{Domain and deployment scope.}
We evaluate only face forgeries; non-facial deepfakes (e.g., full-body synthesis, document forgery) may need domain-specific attribution, such as temporal attention for video. Extending supportability to multi-turn and multilingual explanations also remains open.

\paragraph{Training efficiency.}
TU and VS require extra forward passes, roughly doubling per-step training cost (inference is unaffected). Cached attributions, batch-shared perturbations, or lightweight proxy models could improve scalability.

\end{document}